\documentclass[11pt]{article}

\usepackage[final]{acl}

\usepackage{times}
\usepackage{latexsym}

\usepackage[T1]{fontenc}

\usepackage[utf8]{inputenc}

\usepackage{microtype}

\usepackage{inconsolata}

\usepackage{graphicx}

\newcommand{\logo}{\textsc{PsychePass}\xspace}
\newcommand{\iconopen}{\hspace{0.2em}\includegraphics[height=0.7em]{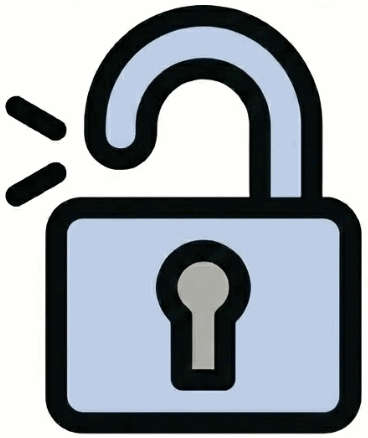}}
\newcommand{\iconclose}{\hspace{0.3em}\includegraphics[height=0.7em]{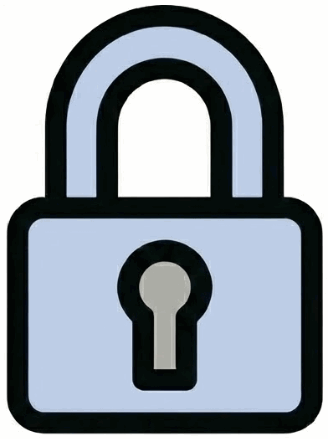}}
\newcommand{\iconpsi}{\hspace{0.3em}\includegraphics[height=0.7em]{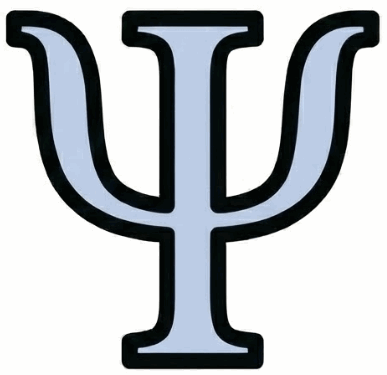}}

\newcommand{\gradientcell}[1]{%
    \begin{tikzpicture}[baseline=0.05cm]
        \definecolor{steelblue}{RGB}{70,130,180}
        \node[anchor=east, text width=1.2em, align=center] at (0,0.15) {\textbf{#1}};
        \fill[steelblue!10] (0.1,0) rectangle (1.1,0.3); 
        \fill[steelblue!80] (0.1,0) rectangle ({0.1+1.0*#1/143},0.3); 
    \end{tikzpicture}%
}

\usepackage{subcaption}
\usepackage{enumitem}
\usepackage{tikz}
\usepackage{multirow}
\usepackage{pifont}
\usepackage{amssymb}
\usepackage{bm}
\usepackage{amsmath}
\usepackage{booktabs}
\usepackage{makecell}
\usepackage{colortbl}
\usepackage{array}
\usepackage{algorithm}
\usepackage{algorithmic}
\usepackage{graphicx}
\usepackage[figuresright]{rotating}
\usepackage{xspace} 
\usepackage{bbold}
\usepackage{tabularx}
\usepackage{fontawesome}
\usepackage{comment}
\usepackage{tcolorbox}

\usepackage{xcolor}
\usepackage{xfp}
\definecolor{JungleGreen}{rgb}{0.16, 0.67, 0.53}

\definecolor{s4blue}{RGB}{135,169,206}
\definecolor{s4pink}{RGB}{255,182,193}
\definecolor{s4gray}{RGB}{128,128,128}

\definecolor{catRel}{RGB}{217, 230, 242}  
\definecolor{catSkill}{RGB}{198, 220, 235}  
\definecolor{catBase}{RGB}{217, 230, 242}  

\definecolor{pinegreen}{RGB}{15,153,15}
\definecolor{mygray}{gray}{.9}
\definecolor{target}{RGB}{0,0,146}
\definecolor{myblue}{RGB}{25,101,255}
\definecolor{myorange}{RGB}{239,134,63}

\newcommand*{\circled}[1]{\lower.7ex\hbox{\tikz\draw (0pt, 0pt)%
		circle (.5em) node {\makebox[1em][c]{\small #1}};}}

\definecolor{bestd}{RGB}{237,100,152}
\definecolor{bestc}{RGB}{0,126,219}

\newcommand{\myroman}[1]{\uppercase\expandafter{\romannumeral#1}}

\definecolor{DeepSteelBlue}{RGB}{135,169,206}

\newcommand{\colgradient}[1]{\cellcolor{DeepSteelBlue!\fpeval{(#1-10)/(208-10)*100}}#1}

\title{\logo: Calibrating LLM Therapeutic Competence\\via Trajectory-Anchored Tournaments}

  \author{
	\textbf{Zhuang Chen\textsuperscript{1}},
	\textbf{Dazhen Wan\textsuperscript{2}},
	\textbf{Zhangkai Zheng\textsuperscript{3}},
	\textbf{Guanqun Bi\textsuperscript{4}},\\
	\textbf{Xiyao Xiao\textsuperscript{2}},
	\textbf{Binghang Li\textsuperscript{2}},
	\textbf{Minlie Huang\textsuperscript{4}}
	\\
	\\
	\textsuperscript{1}School of Computer Science and Engineering, Central South University\\
	\textsuperscript{2}Lingxin AI\quad
	\textsuperscript{3}South China Normal University \\
	\textsuperscript{4}CoAI Group, DCST, IAI, BNRIST, Tsinghua University\\
	{zhchen18@foxmail.com \quad biguanqun@mail.tsinghua.edu.cn}}

\begin{document}
\maketitle
\begin{abstract}
	While large language models show promise in mental healthcare, evaluating their therapeutic competence remains challenging due to the unstructured and longitudinal nature of counseling. We argue that current evaluation paradigms suffer from an \textit{unanchored defect}, leading to two forms of instability: \textit{process drift}, where unsteered client simulation wanders away from specific counseling goals, and \textit{standard drift}, where static pointwise scoring lacks the stability for reliable judgment. To address this, we introduce \logo, a unified framework that calibrates the therapeutic competence of LLMs via trajectory-anchored tournaments. We first anchor the \textit{interaction trajectory} in simulation, where clients precisely control the fluid consultation process to probe multifaceted capabilities. We then anchor the \textit{battle trajectory} in judgments through an efficient Swiss-system tournament, utilizing dynamic pairwise battles to yield robust Elo ratings. Beyond ranking, we demonstrate that tournament trajectories can be transformed into credible reward signals, enabling on-policy reinforcement learning to enhance LLMs' performance. Extensive experiments validate the effectiveness of \logo and its strong consistency with human expert judgments.

\end{abstract}

\section{Introduction}

The explosive growth of large language models offers a beacon of hope for the mental healthcare sector, functioning as scalable, 24/7 counseling assistant support systems, alleviating the heavy burden of professional therapists \cite{stade2024large}. However, unlike deterministic tasks such as code generation, counseling is a highly unstructured, longitudinal process inherently reliant on dynamic interaction and systematic stage progression \cite{norcross2011psychotherapy}. Consequently, accurately evaluating the therapeutic competence of LLMs remains a pressing challenge in the field \cite{na2025survey}.

\begin{figure}[t]
    \setlength{\abovecaptionskip}{1mm}
    \setlength{\belowcaptionskip}{1mm}
    \vspace{0em}
    \centering
    \includegraphics[width=\linewidth]{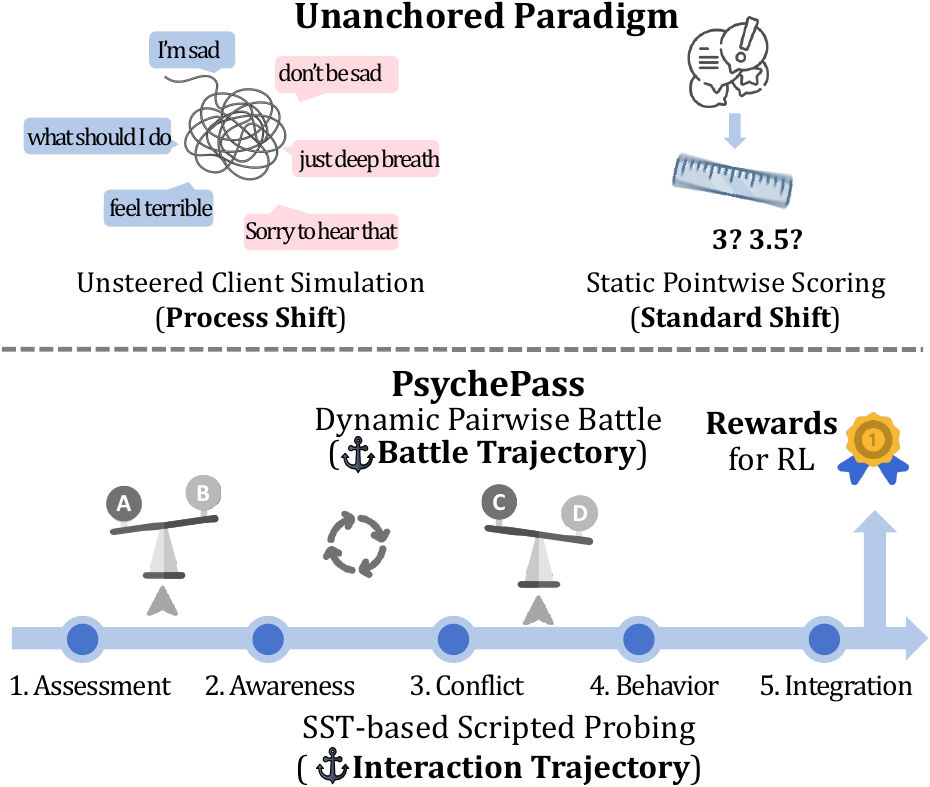}
    \caption{\logo anchors two traceable trajectories for calibrating LLMs' therapeutic competence.}
    \label{fig:intro}
    \vspace{-2em}
\end{figure}

Simulating client-therapist interactions is the natural and effective approach for evaluating counseling competence, mirroring the supervision of human counselors. While a few attempts \cite{zhu2025psi,wang2025care} adopt this paradigm, we identify a fundamental ``\textbf{unanchored defect}'' that limits their effectiveness. This manifests in two dimensions:
1) \textit{Unanchored simulation leads to process drift.} Most evaluations rely on single-turn QA or aimless free-form chats based on static client profiles. These unsteered simulations frequently wander without clinical direction, failing to trigger high-stakes scenarios such as trauma processing or skill application. As a result, evaluations capture only superficial features, such as generic empathy, rather than true therapeutic competence.
2) \textit{Unanchored judgment leads to standard drift.} Existing methods predominantly use static pointwise scoring (e.g., Likert scales), forcing model judges to rely on vague standards without comparative references. This often results in evaluation collapse \cite{wang2025improving}, where scores cluster in a narrow range with poor discriminability. The resulting lack of clear, comparative signals makes it impossible to accurately rank and optimize models.


To overcome the unanchored defect, we introduce \textsc{PsychePass}, a calibration framework for ranking and optimizing the therapeutic competence of LLMs by anchoring two traceable trajectories.

\noindent \textbf{1) Anchoring the interaction trajectory in simulation.} To eliminate process drift and ensure the interactions between simulated clients and evaluated models reach sufficient depth, we script \textit{interaction trajectories} grounded in single-session therapy \cite{talmon1990single} theory. The simulated client drives the dialogue flow through five complete counseling stages, ranging from alliance building to behavior correction, and performs targeted probing of 12 specific competency dimensions, allowing disentangled measurement of multifaceted capabilities.

\noindent \textbf{2) Anchoring the battle trajectory in judgment.} To resolve standard drift and ensure discriminative judgment, we abandon static pointwise scoring. Instead, we construct explicit \textit{battle trajectories} by determining winners through pairwise battles. We employ a Swiss-system tournament \cite{csato2013ranking} to reduce comparison complexity from $N(N-1)/2$ to $(N/2)\log_2 N$, significantly lowering computational costs. Based on battle results, we utilize the Bradley-Terry model \cite{bradley1952rank} to calculate Elo scores \cite{elo1978rating}, yielding stable and comparable rankings.

Beyond evaluation, we further demonstrate that tournament trajectories can be transformed into credible reward signals to support on-policy reinforcement learning to enhance LLMs' therapeutic competence, establishing a complete closed calibration loop from evaluation to optimization. For experiments, we first evaluate the therapeutic competence of $12$ LLMs, including cutting-edge general models and domain-specific models tailored for counseling, and provide fine-grained, discriminative performance profiles. Extensive analysis validates the effectiveness of \logo{}, achieving high inter-rater agreement with psychologists (Cohen's $\kappa > 0.7$). We then demonstrate that, after being trained with trajectory-based RL, a target LLM achieves a win-to-loss ratio of $62:25$ compared to its original counterpart.

Our main contributions are: 1) Introducing \logo to establish the landscape of therapeutic competence for concurrent LLMs; 2) Exploring how to obtain high-fidelity rewards for aligning LLMs' counseling capabilities via on-policy RL; 3) Conducting extensive experiments verifying that \logo's calibration is credible and qualified for serving as a foundation for future work.

\begin{table*}[t]
    \centering
    \footnotesize

    \renewcommand\arraystretch{1.0}
    \setlength{\abovecaptionskip}{1mm}
    \setlength{\belowcaptionskip}{0mm}
    \newcommand{\cempty}{\tikz[baseline=-0.5ex]{\draw[black] (0,0) circle (0.5ex);}\hspace{0.3em}}
    \newcommand{\chalf}{\tikz[baseline=-0.5ex]{\draw[black] (0,0) circle (0.5ex); \fill[black] (90:0.5ex) arc (90:270:0.5ex) -- cycle;}\hspace{0.3em}}
    \newcommand{\cfull}{\tikz[baseline=-0.5ex]{\fill[black] (0,0) circle (0.5ex);}\hspace{0.3em}}
    \begin{tabular}{p{5cm}p{3cm}p{3cm}p{3cm}}
        \toprule
        \multicolumn{1}{l}{\textbf{Systems}}           & \multicolumn{1}{l}{\textbf{Simulation Paradigm}} & \multicolumn{1}{l}{\textbf{Judgement Paradigm}} & \multicolumn{1}{l}{\textbf{Optimization Loop}} \\
        \midrule
        Psyche-R1 \cite{dai2025psyche}                 & \cempty Knowledge QA                             & \cempty Accuracy                                & \chalf Knowledge RL                            \\
        MentraSuite \cite{xiao2025mentrasuite}         & \cempty Knowledge QA                             & \cempty Accuracy                                & \chalf Knowledge SFT-RL                        \\
        PsychCounsel-Bench \cite{zeng2025psychcounsel} & \cempty Knowledge QA                             & \cempty Accuracy                                & \cempty      Eval Only                         \\
        CBT-Bench \cite{zhang2025cbt}                  & \cempty Knowledge QA                             & \cempty Accuracy                                & \cempty    Eval Only                           \\
        CounselBench \cite{li2025counselbench}         & \cempty Knowledge QA                             & \cempty Pointwise                               & \cempty      Eval Only                         \\
        ClientCAST \cite{wang2024towards}              & \cempty Free-form Chat                           & \cempty Pointwise                               & \cempty    Eval Only                           \\
        MindEval \cite{pombal2025mindeval}             & \cempty Free-form Chat                           & \cempty Pointwise                               & \cempty Eval Only                              \\
        CARE-Bench \cite{wang2025care}                 & \chalf Guided Chat                               & \cempty Pointwise                               & \cempty
        Eval Only                                                                                                                                                                                            \\
        $\psi$-Arena \cite{zhu2025psi}                 & \chalf Guided Chat                               & \cempty Pointwise                               & \chalf Post-hoc Patch                          \\
        \midrule
        \rowcolor{gray!10} \textbf{\logo{} (Ours)}     & \textbf{\cfull Scripted Probing}                 & \textbf{\cfull Pairwise}                        & \textbf{\cfull Trajectory RL}                  \\
        \bottomrule
    \end{tabular}
    \caption{Comparison of different studies in evaluating therapeutic competencies of LLMs.}
    \label{tab:comparison}
    \vspace{-1em}
\end{table*}



\section{Related Work}
\label{sec:related_work}

Research on LLMs for psychological counseling has expanded rapidly, ranging from synthesizing counseling dialogue corpora \cite{zhang2024cpsycoun,DBLP:conf/acl/XieCXLX25} to incentivizing diagnostic reasoning \cite{DBLP:journals/tcss/HuDLMZSGYW25, DBLP:journals/corr/abs-2505-15715,dai2025psyche,xiao2025mentrasuite}. However, the real-world counseling process is inherently unstructured and longitudinal, relying on the sophisticated orchestration of diverse therapeutic competencies. Consequently, while the pathway to training conversational agents that can generate superficially plausible counseling dialogues is becoming clear, the rigorous evaluation of their actual clinical effectiveness lags significantly behind their development.

\noindent\textbf{Static Knowledge Examination} Current assessments predominantly rely on static benchmarks. PsychCounsel-Bench \cite{zeng2025psychcounsel} and CBT-Bench \cite{zhang2025cbt} utilize professional exams or fixed case studies, while CounselBench \cite{li2025counselbench} focuses on case Q\&As. While these benchmarks effectively measure theoretical knowledge, they fall short of verifying whether a model can handle the unstructured, longitudinal dynamics of a real counseling session.

\noindent\textbf{Interactive Simulation Tests} To move beyond static tests, recent studies develop interactive simulations where LLMs chat with simulated clients \cite{pombal2025mindeval,wang2024towards}. However, these environments primarily rely on free-form chats and pointwise scoring, which are prone to process and standard drift. CARE-Bench \cite{wang2025care} partially guides client simulation with expert principles but lacks explicit stage-level control. $\Psi$-Arena \cite{zhu2025psi} simply incorporates counseling stages in prompting yet does not enforce strict process adherence. Moreover, only a few studies incorporate optimization beyond evaluation. Psyche-R1 \cite{dai2025psyche} and MentraSuite \cite{xiao2025mentrasuite} optimize knowledge reasoning through RL, but their objectives focus on answering exam questions. $\Psi$-Arena employs post-hoc prompt patching for response revision without internalizing the counseling capabilities.

In contrast, \logo{} constrains simulations to explicit probing paths and adopts pairwise comparisons to eliminate process and standard drift, and further establishes a closed loop for evaluation and trajectory-level optimization. We present a straightforward comparison between the proposed \logo{} and existing methods in Table \ref{tab:comparison}.

\section{\logo}
\label{sec:method}

\begin{figure*}[t]
    \setlength{\abovecaptionskip}{1mm}
    \setlength{\belowcaptionskip}{1mm}
    \centering
    \includegraphics[width=\textwidth]{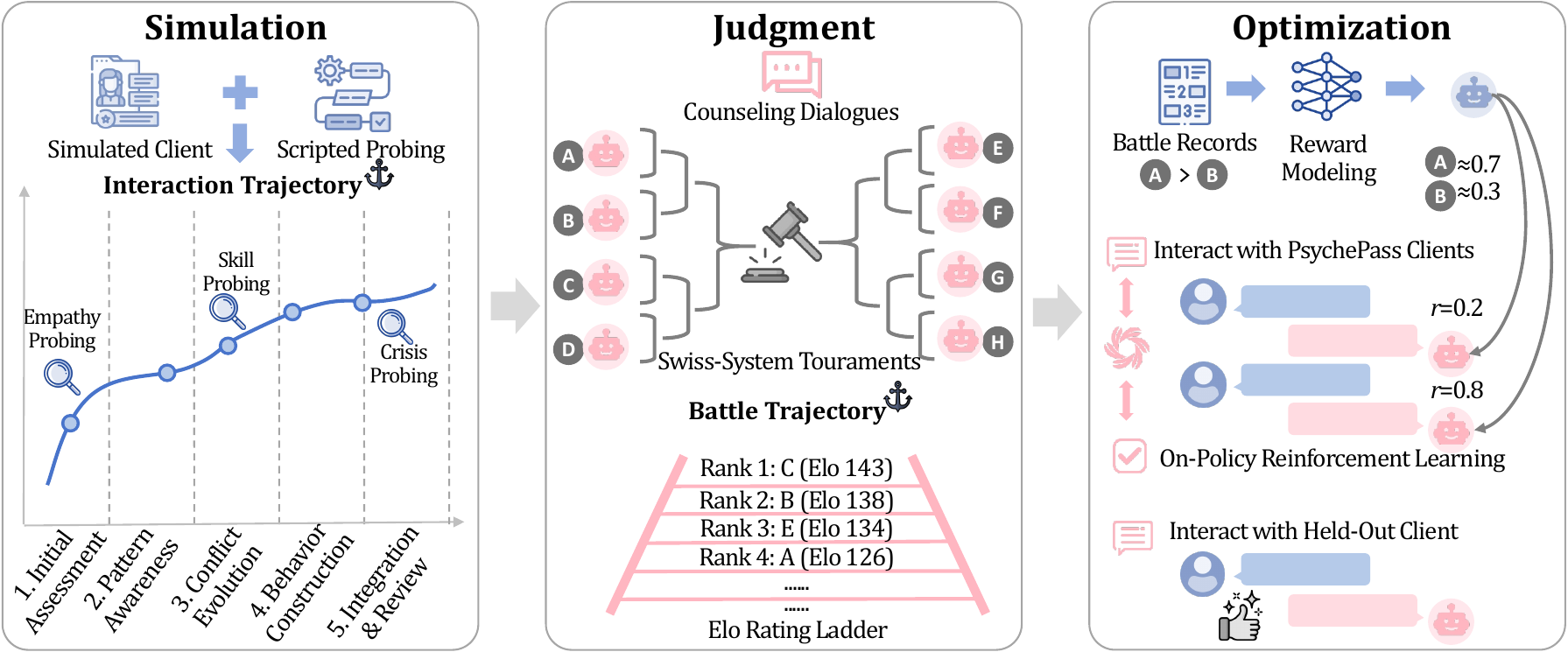}
    \caption{The overall framework of \logo.}
    \label{fig:overview}
    \vspace{-1.5em}
\end{figure*}

\logo calibrates the therapeutic competence of LLMs in three steps (see Figure \ref{fig:overview}): \textbf{1) Simulation}: Each model interact with scripted clients to generate counseling conversations. \textbf{2) Judgment}: All models participate in a Swiss-system tournament and battle with each other to get Elo ratings. \textbf{3) Optimization}: The tournament results are transformed into reward signals to optimize models via on-policy RL and boost their counseling performance in unseen scenarios.

\subsection{Interaction Trajectory in Simulation}
\label{sec:interaction_trajectory}

In authentic clinical settings, counseling is not a random walk but a structured journey driven by specific therapeutic goals. To replicate this, we define ``\textit{interaction trajectory}'' as the conversational path between therapists and clients, and anchor it by transforming the path from unstructured, open-ended chats into rigorous, scripted probing. Simulated clients function as examiners, explicitly assessing specific therapeutic competencies at designated positions within the counseling process, forming a ``stress test''.

\subsubsection{Client Profiles}

A valid counseling trajectory necessitates a realistic starting point. We initialize client profiles using seed data from a real-world online counseling platform YiSum\footnote{\url{https://www.xinli001.com/}. We have obtained official permission to use the anonymized data from the platform.}, which are then de-identified and expanded by LLMs into high-fidelity personas. To capture the nuance of real clinical cases, we define a comprehensive schema for each profile. Beyond basic demographics, we incorporate deep psychological attributes: \texttt{Core Drive}, representing the fundamental need or fear motivating behavior; \texttt{Reaction Pattern}, defining typical defense mechanisms under stress; and \texttt{Situation \& Event}, a detailed reconstruction of the core conflict including participants, locations, and emotional triggers. We build an initial pool of over $10,000$ candidate profiles, and sort each profile based on its linguistic complexity, emotional intensity, and counseling relevance. Finally, we select $100$ high-quality profiles to serve as simulated clients, each verified by human experts for anonymity and harmlessness. We use a LLM $\mathcal{M}_{client}$ to role-play as clients. Note that each client supporting a counseling session of $40$--$50$ turns, this yields approximately $4,000$--$5,000$ conversation turns per therapist, providing a substantial foundation for evaluation. {Details of client profiles can be found in Appendix A.}


\subsubsection{Target Competency Dimensions}

To align our evaluation with professional standards, we initially reference competency frameworks from the American Psychiatric Association (APA) \cite{kupersanin2002apa}. However, direct application of human standards to LLMs is insufficient; some metrics are inapplicable to text-based agents (e.g., observation of non-verbal cues), while models face unique challenges such as maintaining long-term memory and response diversity.
We address this gap via a human-in-the-loop pilot experiment. We deploy several LLMs selected from the evaluation pool and invite approximately 200 participants to engage in month-long counseling sessions, providing real-time feedback on responses. We then analyze these annotated interactions to extract latent criteria, which are then screened and consolidated by human experts.

This process yields 12 distinct competency dimensions, organized into three categories: \textbf{Alliance Building}, including \texttt{Empathy}, \texttt{Discernment}, \texttt{Engagement}; \textbf{Professional Technique}, including \texttt{Skill}, \texttt{Suggestion}, \texttt{Reframing}, \texttt{Progression}, \texttt{Trauma}; and \textbf{Reliability Support}, including \texttt{Crisis}, \texttt{Ethics}, \texttt{Diversity}, \texttt{Memory}. Among these, \texttt{Empathy}, \texttt{Engagement}, and \texttt{Diversity} serve as {global metrics}, evaluated holistically at the session level (e.g., we use Distinct-N to evaluate \texttt{Diversity}). The remaining nine function as {local metrics}, where the simulated client triggers specific responses at designated positions to enable targeted assessment. We design delicate probing methods for these local dimensions. For instance, regarding \texttt{Skill}, the client might say \textit{``I wish I could talk to my father face-to-face, but he passed away''} to test if the model applies the empty chair technique \cite{perls1951gestalt}. Similarly, addressing the common hallucination issue in LLMs, we construct memory-specific questions (e.g., \textit{``Remember that conflict with my brother?''}) to rigorously test \texttt{Memory}. Note that although counseling is primarily client-driven, we also evaluate the model's ability to drive the conversation forward (i.e., \texttt{Progression} dimension). We create opportunities by leaving several turns empty without any driving after a phase clearly concludes, observing whether the model proactively advances to the next stage. {Detailed definitions, probing methods, and examples are provided in Appendix \ref{app:competency_dimensions}.}

\subsubsection{Interaction Trajectory Anchoring}

To prevent simulated dialogues from collapsing into superficial interactions (e.g., merely showing empathy) and to ensure a comprehensive assessment of therapeutic competencies, we anchor the simulation to a rigorous scripted probing process grounded in single-session therapy (SST). We structure the dialogue into five distinct phases, denoted as $\Phi = \{\phi_1, \dots, \phi_5\}$, corresponding to {alliance building \& initial assessment}, {pattern awareness \& issue concretization}, {core conflict evolution \& trauma processing}, {corrective experience \& new behavior construction}, and {integration, review \& termination} in SST, respectively.
By integrating these phases with our competency framework, we generate a detailed execution script for each client. This script strictly adheres to the progression of $\Phi$ and strategically triggers responses targeting the nine local competency dimensions $\mathcal{D} = \{d_1, \dots, d_9\}$ at clinically appropriate junctures. Consequently, at each turn $t$, we dynamically update the system prompt of clients $\mathcal{M}_{client}$ following this trajectory:
\begin{equation}
    \small
    \setlength{\abovedisplayskip}{2pt}
    \setlength{\belowdisplayskip}{2pt}
    \mathcal{P}_{sys}^{(t)} = \Psi(\mathbf{C}, \phi_t, d_t, t)
\end{equation}
where $\mathcal{P}_{sys}^{(t)}$ is the updated prompt, controlled by the client profile $\mathbf{C}$, the current clinical phase $\phi_t \in \Phi$, the target competency dimension $d_t \in \mathcal{D}$, and the dialogue turn $t$. This mechanism enables the client to precisely probe specific dimensions by triggering corresponding responses, thereby anchoring the interaction trajectory to ensure comprehensive coverage of all competency dimensions throughout the simulation. {An example of the execution script is provided in Appendix C.}

\subsection{Battle Trajectory in Judgment}

To overcome the unstable outcomes of static pointwise scoring, we anchor judgment standards through dynamic pairwise battles. We define ``\textit{battle trajectory}'' as the evolving ranking of models derived from iterative competitions. Unlike absolute scores which suffer from poor discriminability, the battle trajectory establishes relative and robust anchors by determining winners in direct contests.

\subsubsection{Multifaceted Pairwise Comparison}

The basic unit of our evaluation is a {pairwise battle} between two models. Each competence dimension is evaluated according to its granularity. {Global dimensions} (e.g., \texttt{Empathy}) are evaluated by holistically contrasting full sessions from model A and B. {Local dimensions} (e.g., \texttt{Skill}) are assessed only at specific turns triggered by the simulation script. For each battle, the comparison results of all dimensions are judged simultaneously in a single process. The partial orders of battles drive the subsequent Elo ranking, and also serve as the signals for training reward models.

\subsubsection{Swiss-System Tournament}

A full round-robin tournament scales quadratically ($O(N^2)$), becoming prohibitively expensive for large model pools. We address this by implementing a four-round Swiss-system tournament. In each round, models are matched with opponents having similar win records, ensuring that strong models face each other while weaker ones compete amongst themselves. This mechanism efficiently sorts competitors with a complexity of $O(N \log N)$. Our experiments demonstrate that the Swiss-system generates rankings highly consistent with full round-robin results while approximately doubling the efficiency.

We also address the substantial position bias observed in LLM judges, where the order of presentation significantly influences outcomes. Pilot experiments indicate that simple position swapping is insufficient to eliminate this bias, which directly compromises the quality of pairwise comparisons. To resolve this, we employ a \textit{stage slicing} debiasing strategy. Specifically, we segment the dialogue into coherent stages $\{s_1, s_2, \dots, s_n\}$ and present them in an alternating pattern: model A's $s_1$, B's $s_1$, B's $s_2$, A's $s_2$, and so forth. This cross-positioned presentation cancels out the first-mover advantage across stages and reduces position bias to a negligible level. We validate the effectiveness of this debiasing approach in experiments.

\subsubsection{Elo Rating Estimation}

We employ the Bradley-Terry (BT) model to convert pairwise outcomes into stable scalar metrics. Unlike traditional Elo systems that update sequentially and are sensitive to match order, we treat Elo ratings as static parameters to be optimized globally. We model the probability that model A defeats B as a sigmoid function of their rating difference. We estimate ratings using maximum likelihood estimation to fit the observed tournament outcomes, formalized in Algorithm \ref{alg:bt_elo}. In \logo, we set the baseline rating to $100$ and the scaling factor $\xi = 400 / \ln(10)$. This ensures that the entire judgment process is referenced and traceable, effectively anchoring the battle trajectory for valid and robust assessment.

\begin{algorithm}[h]
    \caption{\small Bradley-Terry Elo Rating Estimation}
    \label{alg:bt_elo}
    \small
    \begin{algorithmic}[1]
        \REQUIRE Battle records $\mathcal{D} = \{(m_A, m_B, y) \mid y \in \{0, 0.5, 1\}\}$, Learning rate $\eta$
        \ENSURE Final Elo ratings $\mathbf{r}^*$
        \STATE Initialize ratings $\mathbf{r}$ for all models to baseline
        \REPEAT
        \STATE Compute expected win probability for each pair in $\mathcal{D}$:
        \STATE $P(A \succ B) = \frac{1}{1 + e^{-(r_A - r_B) / \xi}}$
        \STATE Compute Negative Log-Likelihood Loss:
        \STATE $\mathcal{L}(\mathbf{r}) = - \sum_{(A,B,y) \in \mathcal{D}} [y \log P(A \succ B) + (1-y) \log (1 - P(A \succ B))]$
        \STATE Update ratings via Gradient Descent:
        \STATE $\mathbf{r} \leftarrow \mathbf{r} - \eta \nabla \mathcal{L}(\mathbf{r})$
        \UNTIL{convergence}
        \RETURN $\mathbf{r}$
    \end{algorithmic}
\end{algorithm}
\vspace{-1em}

\subsection{Trajectory-based RL for Optimization}
\label{sec:alignment}

Beyond evaluation, we investigate whether high-fidelity trajectory data can actively enhance therapeutic capabilities. We propose trajectory-based reinforcement learning that converts evaluative judgments into actionable training signals.

We first train a general reward model $\mathcal{M}_{RM}$ capable of distinguishing therapeutic quality across different dimensions. The training data is derived from the partial orders of battles observed in the tournaments. Specifically, for a target competency dimension $d$, if response $o_w$ is judged better than $o_l$, we optimize $\mathcal{M}_{RM}$ to assign a higher scalar reward to $o_w$:
\begin{equation}
    \small
    \setlength{\abovedisplayskip}{2pt}
    \setlength{\belowdisplayskip}{2pt}
    \mathcal{L}_{RM} = - \mathbb{E}_{(d, o_w, o_l) \sim \mathcal{T}} \bigg[ \log \sigma \Big( r(o_w, d) - r(o_l, d) \Big) \bigg]
\end{equation}
where $\sigma$ is the sigmoid function and $r(\cdot)$ represents the reward score predicted by $\mathcal{M}_{RM}$ given the context and specific dimension.

We then select a target model for optimization. To prevent distribution shift, we conduct on-policy training using group relative policy optimization \cite{shao2024deepseekmath} on the same 100 client profiles used in the tournaments. For each client query $q$, we sample a group of responses $\{o_i\}_{i=1}^G$ and compute their dimension-specific rewards using $\mathcal{M}_{RM}$. The objective is:
\begin{equation}
    \small
    \setlength{\abovedisplayskip}{2pt}
    \setlength{\belowdisplayskip}{2pt}
    \begin{aligned}
        \mathcal{J}_{GRPO}(\theta)              & = \mathbb{E}_{q \sim P(Q), \{o_i\} \sim \pi_{\theta_{old}}} \bigg[ \frac{1}{G} \sum_{i=1}^G \Big( \mathcal{L}^{clip}_i \\
                                                & \quad - \beta \mathbb{D}_{KL}(\pi_{\theta} || \pi_{ref}) \Big) \bigg],                                                 \\
        \text{where} \quad \mathcal{L}^{clip}_i & = \min \left( \rho_i A_i, \text{clip}(\rho_i, 1-\epsilon, 1+\epsilon) A_i \right).
    \end{aligned}
\end{equation}
The advantage $A_i= (r_i - \mu_G) / \sigma_G$, where $r_i$ is the reward from $\mathcal{M}_{RM}$ for response $o_i$, and $\mu_G, \sigma_G$ are the mean and standard deviation across the $G$ responses. The term $\rho_i = \pi_{\theta}(o_i|q) / \pi_{\theta_{old}}(o_i|q)$ represents the importance weight, while $\beta$ controls the KL divergence penalty.

A critical risk here is that the trained model might merely memorize the interaction patterns of high-scoring responses rather than learning generalized therapeutic principles. Therefore, we collect an additional set of 100 client profiles as a held-out test set. These clients do not overlap with the optimization set. We compare the performance of the model before and after RL on this independent test set to strictly verify its generalizability.

\section{Experiments}
\label{sec:experiments}

\begin{table*}[t]
  \centering
  \setlength{\tabcolsep}{1mm}
  \footnotesize

  \setlength{\abovecaptionskip}{1mm}
  \setlength{\belowcaptionskip}{1mm}
  \renewcommand\arraystretch{1.0}

  \resizebox{0.99\textwidth}{!}{
    \begin{tabular}{llcccccccccccccc}
      \toprule
      \multicolumn{1}{c}{\multirow{2}{*}{\textbf{Rank}}}          & \multirow{2}{*}{\textbf{Model}} & \multirow{2}{*}{\textbf{Elo Score}} & \multicolumn{1}{c}{\multirow{2}{*}{\textbf{Win Rate}}} & \multicolumn{3}{c}{\textbf{Alliance}} & \multicolumn{5}{c}{\textbf{Technique}} & \multicolumn{4}{c}{\textbf{Reliability}}                                                                                                                                                                                     \\
      \cmidrule(lr){5-7} \cmidrule(lr){8-12} \cmidrule(lr){13-16} &                                 &                                     &                                                        & {\texttt{Emp.}}                       & {\texttt{Dsc.}}                        & {\texttt{Eng.}}                          & {\texttt{Skl.}}   & {\texttt{Sug.}}   & {\texttt{Rfm.}}   & {\texttt{Prg.}}   & {\texttt{Trm.}}   & {\texttt{Crs.}}   & {\texttt{Eth.}}   & {\texttt{Div.}}   & {\texttt{Mem.}}   \\
      \midrule
      \# 1                                                        & GPT-5.2\iconclose               & \gradientcell{143}                  & 82.30\%                                                & \colgradient{131}                     & \colgradient{148}                      & \colgradient{180}                        & \colgradient{166} & \colgradient{208} & \colgradient{150} & \colgradient{191} & \colgradient{171} & \colgradient{171} & \colgradient{155} & \colgradient{134} & \colgradient{149} \\
      \# 2                                                        & Gemini 3 Pro\iconclose          & \gradientcell{138}                  & 75.48\%                                                & \colgradient{163}                     & \colgradient{157}                      & \colgradient{156}                        & \colgradient{165} & \colgradient{146} & \colgradient{163} & \colgradient{163} & \colgradient{166} & \colgradient{116} & \colgradient{142} & \colgradient{148} & \colgradient{146} \\
      \# 3                                                        & Claude Opus 4.5\iconclose       & \gradientcell{134}                  & 75.01\%                                                & \colgradient{144}                     & \colgradient{154}                      & \colgradient{165}                        & \colgradient{159} & \colgradient{127} & \colgradient{141} & \colgradient{165} & \colgradient{164} & \colgradient{144} & \colgradient{150} & \colgradient{92}  & \colgradient{147} \\
      \# 4                                                        & Qwen3-235B-A22B\iconopen        & \gradientcell{126}                  & 69.67\%                                                & \colgradient{153}                     & \colgradient{141}                      & \colgradient{129}                        & \colgradient{144} & \colgradient{126} & \colgradient{142} & \colgradient{135} & \colgradient{146} & \colgradient{103} & \colgradient{128} & \colgradient{133} & \colgradient{136} \\
      \# 5                                                        & GLM-4.6\iconopen                & \gradientcell{114}                  & 55.27\%                                                & \colgradient{130}                     & \colgradient{119}                      & \colgradient{105}                        & \colgradient{117} & \colgradient{99}  & \colgradient{114} & \colgradient{105} & \colgradient{118} & \colgradient{102} & \colgradient{109} & \colgradient{178} & \colgradient{118} \\
      \# 6                                                        & DeepSeek-V3.2\iconopen          & \gradientcell{104}                  & 50.36\%                                                & \colgradient{98}                      & \colgradient{95}                       & \colgradient{95}                         & \colgradient{93}  & \colgradient{96}  & \colgradient{95}  & \colgradient{93}  & \colgradient{95}  & \colgradient{106} & \colgradient{107} & \colgradient{180} & \colgradient{100} \\
      \# 7                                                        & PsyLLM\iconpsi                  & \gradientcell{90}                   & 46.24\%                                                & \colgradient{139}                     & \colgradient{115}                      & \colgradient{103}                        & \colgradient{93}  & \colgradient{79}  & \colgradient{104} & \colgradient{73}  & \colgradient{82}  & \colgradient{74}  & \colgradient{70}  & \colgradient{23}  & \colgradient{81}  \\
      \# 8                                                        & SoulChat2.0\iconpsi             & \gradientcell{89}                   & 45.62\%                                                & \colgradient{86}                      & \colgradient{86}                       & \colgradient{87}                         & \colgradient{80}  & \colgradient{81}  & \colgradient{86}  & \colgradient{80}  & \colgradient{76}  & \colgradient{82}  & \colgradient{79}  & \colgradient{94}  & \colgradient{85}  \\
      \# 9                                                        & Crispers-14B\iconpsi            & \gradientcell{85}                   & 43.65\%                                                & \colgradient{71}                      & \colgradient{73}                       & \colgradient{73}                         & \colgradient{71}  & \colgradient{79}  & \colgradient{77}  & \colgradient{71}  & \colgradient{67}  & \colgradient{79}  & \colgradient{78}  & \colgradient{126} & \colgradient{77}  \\
      \# 10                                                       & PsycoLLM\iconpsi                & \gradientcell{67}                   & 26.81\%                                                & \colgradient{36}                      & \colgradient{44}                       & \colgradient{42}                         & \colgradient{44}  & \colgradient{71}  & \colgradient{54}  & \colgradient{52}  & \colgradient{46}  & \colgradient{82}  & \colgradient{71}  & \colgradient{40}  & \colgradient{60}  \\
      \# 11                                                       & Psyche-R1\iconpsi               & \gradientcell{61}                   & 20.93\%                                                & \colgradient{34}                      & \colgradient{40}                       & \colgradient{34}                         & \colgradient{40}  & \colgradient{54}  & \colgradient{46}  & \colgradient{40}  & \colgradient{45}  & \colgradient{78}  & \colgradient{62}  & \colgradient{25}  & \colgradient{54}  \\
      \# 12                                                       & EmoLLM V3.0\iconpsi             & \gradientcell{44}                   & 4.77\%                                                 & \colgradient{10}                      & \colgradient{21}                       & \colgradient{24}                         & \colgradient{21}  & \colgradient{27}  & \colgradient{22}  & \colgradient{26}  & \colgradient{20}  & \colgradient{57}  & \colgradient{44}  & \colgradient{22}  & \colgradient{42}  \\
      \bottomrule
    \end{tabular}
  }
  \caption{Leaderboard of therapeutic competence with fine-grained Elo scores of $12$ LLMs.}
  \label{tab:main_results}
  \vspace{-1.5em}
\end{table*}

\subsection{Setup}

\noindent \textbf{Simulation} Clients are driven by Doubao-1-5-Pro-32k-Character-250228 \cite{doubao15pro32kcharacter250228} due to its specialized optimization in role-playing. The simulation involves $100$ distinct client profiles. Each test model engages in a full-length ($40$--$50$ turns) counseling session with $100$ clients, resulting in a total of $4,400$ turns per model.

\vspace{0.1em}
\noindent \textbf{Judgment} Twelve competitive models engage in a Swiss-system tournament which consists of $4$ rounds. In each round, models are paired into $6$ matches based on their current standings. To eliminate position bias, each match is played twice with roles swapped, resulting in $12$ individual battles per round per case. Across $100$ clients and $4$ rounds, there are $4,800$ pairwise sessions and $57,600$ battles for 12 competence dimensions. We use DeepSeek-R1 \cite{guo2025deepseek} for judging across all dimensions simultaneously.

\vspace{0.1em}
\noindent \textbf{Optimization} The reward model $\mathcal{M}_{rm}$ is trained based on Qwen3-8B-Instruct \cite{qwen3technicalreport}, the optimizing target $\mathcal{M}_{base}$ is also another Qwen3-8B-Instruct model. Details of prompting and training implementation can be found in Appendix~\ref{app:implmentation_detals}.

\subsection{Models}
To ensure a comprehensive evaluation, we curate a diverse roster of competitors spanning both domain-specific and cutting-edge general LLMs. The specialized psychological group includes SoulChat2.0 \cite{DBLP:conf/acl/XieCXLX25}, PsycoLLM \cite{DBLP:journals/tcss/HuDLMZSGYW25}, EmoLLM\footnote{https://github.com/SmartFlowAI/EmoLLM}(latest ver.3.0, also known as CPsyCounX \cite{zhang2024cpsycoun}), Psyche-R1 \cite{dai2025psyche}, PsyLLM \cite{DBLP:journals/corr/abs-2505-15715}, and Crisper-14B \cite{DBLP:journals/corr/abs-2504-17238}. General LLMs include the open-source Qwen3-235B-A22B-Instruct-2507 \cite{qwen3technicalreport}, GLM-4.6 \cite{zeng2025glm}, and DeepSeek-V3.2 \cite{deepseekai2025deepseekv32}, as well as the proprietary Gemini-3-Pro \cite{google2025gemini}, Claude-Opus-4.5 \cite{anthropic2025claude}, and GPT-5.2 \cite{openai2025gpt52}.

\subsection{Main Results}
Table~\ref{tab:main_results} presents the leaderboard of LLM therapeutic competence\footnote{We have also prepared the online battleground for future real-time competition, see Appendix \ref{sec:app_online_leaderboard}.}. We now dissect the results and highlight several critical observations.

\vspace{0.1em}
\noindent \textbf{Landscape of LLM Competence.} The top tier is exclusively occupied by proprietary frontier general models like GPT-5.2, which achieve a substantial performance lead over all counseling-specific models. This finding contradicts prior studies that often report domain models as competitive or superior. The discrepancy arises because domain models typically emphasize narrow improvements in specific aspects such as empathy or reframing, and may outperform general models on corresponding isolated metrics. However, in rigorous, realistic counseling scenarios, their limited reasoning, world knowledge, and instruction-following capabilities severely constrain their overall competence.

\vspace{0.1em}
\noindent \textbf{Differentiated Dimensional Performance.} Models exhibit distinct profiles across dimensions. For instance, DeepSeek-V3.2 scores remarkably high in \texttt{Diversity} (180), whereas Claude Opus 4.5, despite its high overall rank, scores notably lower (92). We also observe that certain dimensions align closely with the overall ranking, particularly \texttt{Skill} and \texttt{Trauma}, precisely the capabilities that drive genuine client transformation. This pattern underscores the necessity of multi-dimensional evaluation, of which the discriminative power of each dimension will be verified in Section~\ref{sec:analysis}.

\vspace{0.1em}
\noindent \textbf{Opportunities for Domain Models.} Domain models demonstrate competitive performance in empathy. For instance, PsyLLM achieves an \texttt{Empathy} score of 139, even surpassing GPT-5.2. However, they struggle in other dimensions such as \texttt{Diversity}, \texttt{Memory}, and \texttt{Progression}, revealing weak capability in varied response generation and dialogue advancement. These patterns suggest two straightforward directions for developing stronger domain models: first, synthesizing training data that covers diverse scenarios and multiple competencies to ensure comprehensive skill acquisition during SFT; second, adopting multi-agent frameworks that leverage the power of general LLMs to build customized psychological applications.

\subsection{Consistency with Human Experts}

To validate the alignment between \logo and human judgment, we select $30$ cases with battle records and invite two professional counselors with over $1,000$ hours of experience to evaluate the outcomes. Each case is assessed across random $4$ dimensions, yielding $120$ data points for reliable Cohen's kappa calculation. Cases are categorized into three difficulty levels: strong-strong (random $2$ from top $6$), weak-weak (random $2$ from bottom $6$), and strong-weak (top $6$ vs. bottom $6$).

\begin{table}[h]
  \vspace{0em}
  \centering
  \setlength{\abovecaptionskip}{1mm}
  \setlength{\belowcaptionskip}{1mm}
  \footnotesize
  \resizebox{0.48\textwidth}{!}{
    \begin{tabular}{lccc}
      \toprule
      Case Type         & (E1, P) & (E2, P) & (E1, E2) \\
      \midrule
      Strong vs. Strong & 0.576   & 0.569   & 0.845    \\
      Weak vs. Weak     & 0.748   & 0.765   & 0.879    \\
      Strong vs. Weak   & 1.000   & 0.959   & 0.959    \\
      \bottomrule
    \end{tabular}
  }
  \caption{Cohen's kappa between \logo (P) and human experts (E1, E2) across different difficulty levels.}
  \label{tab:human_consistency}
  \vspace{0em}
\end{table}

Table~\ref{tab:human_consistency} shows that \logo achieves near-perfect agreement with human experts in strong-weak matchups, reaching $1.000$ and even surpassing inter-expert consistency. The weak-weak cases also demonstrate substantial agreement around $0.75$. For strong-strong cases, the consistency is relatively lower, which is expected given that experts also exhibit their lowest inter-rater agreement in this most challenging category where performance differences are subtle and hard to distinguish.

\subsection{Effectiveness of Trajectory-Based RL}
\label{sec:rl_impact}

\begin{figure}[h]
  \vspace{-0em}
  \centering

  \setlength{\abovecaptionskip}{1mm}
  \setlength{\belowcaptionskip}{1mm}
  \includegraphics[width=0.48\textwidth]{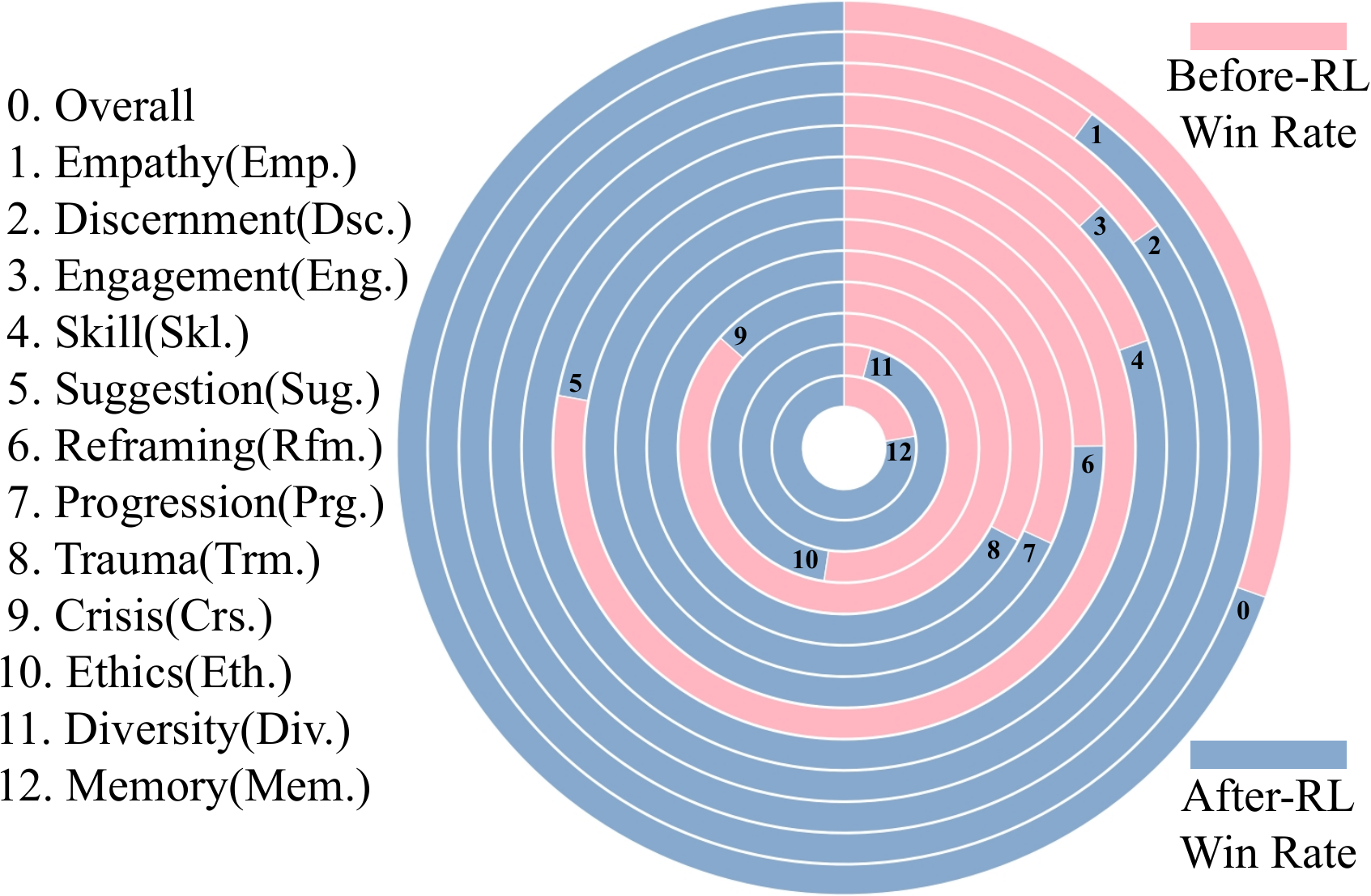}
  \caption{The win rates of the base model and the aligned model across 12 dimensions.}
  \label{fig:rl_comparison}
  \vspace{-0em}
\end{figure}

Figure~\ref{fig:rl_comparison} illustrates the performance shift from the base model ($\mathcal{M}_{base}$) to the RL-aligned model across 12 dimensions. The RL-aligned model achieves an overall win:loss:tie ratio of 59.5:26.0:14.5 in direct battles, demonstrating significant improvements across 9 dimensions. Notably, dimensions such as \texttt{Diversity} (4.0\% $\rightarrow$ 92.0\%) and \texttt{Empathy} (31.0\% $\rightarrow$ 82.0\%) see dramatic gains, confirming that RL effectively instills the core values of client-centered therapy.

Meanwhile, we observe performance regressions in three dimensions: \texttt{Crisis}, \texttt{Suggestion}, and \texttt{Ethics}. These declines are explainable, as general models undergo strong compliance alignment and tend to provide direct advice. The RL training for counseling may reduces this behavior, but remains effective for therapeutic purposes.

\section{Analysis}
\label{sec:analysis}

\subsection{Swiss System Convergence}
\begin{figure}[h]
  \vspace{0em}
  \centering
  \setlength{\abovecaptionskip}{1mm}
  \setlength{\belowcaptionskip}{1mm}
  \hspace{-1em}
  \includegraphics[width=0.46\textwidth]{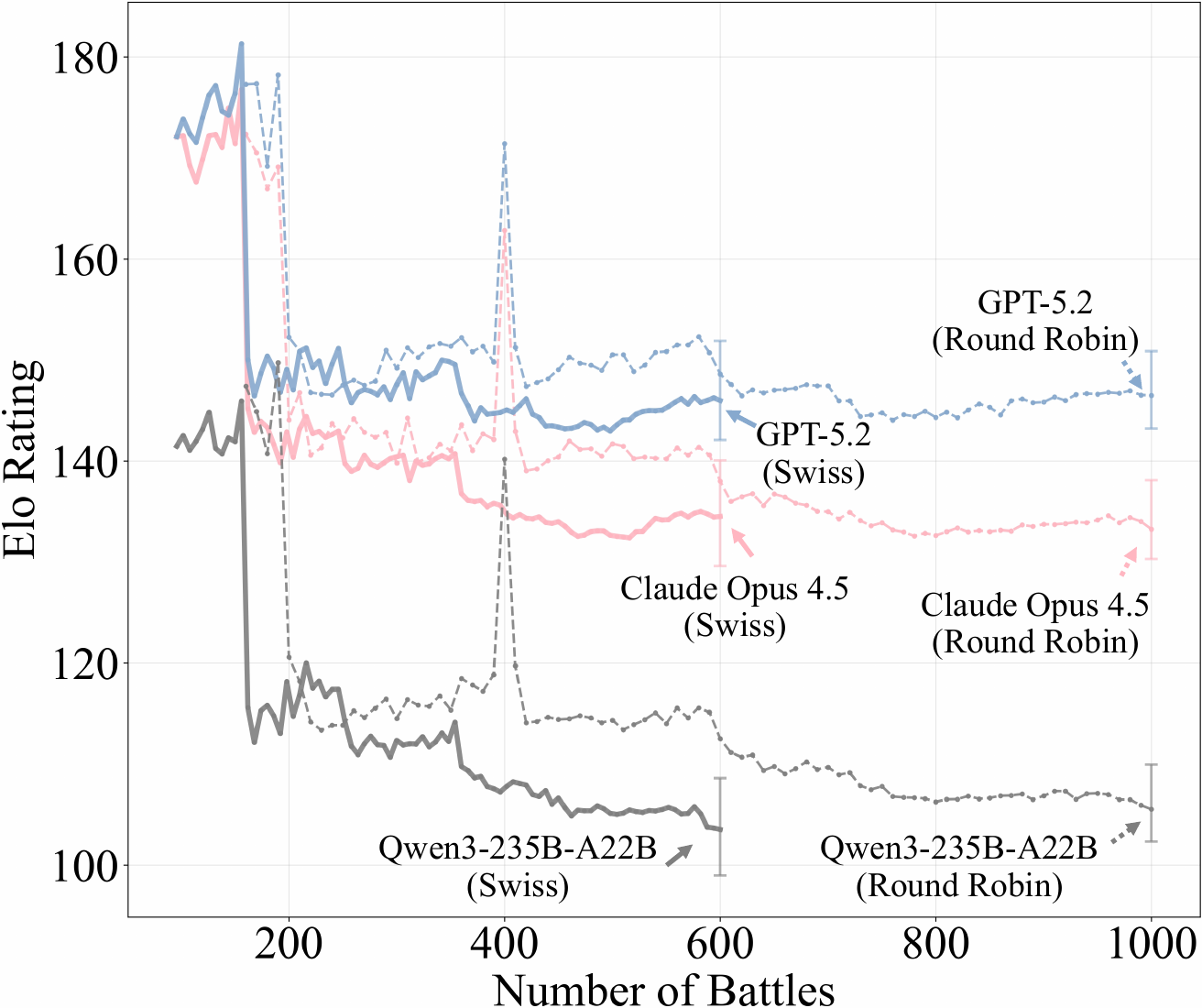}
  \caption{Convergence of Elo ratings.}
  \label{fig:swiss_convergence}
  \vspace{-0.5em}
\end{figure}

To validate the effectiveness and efficiency of the Swiss-system tournament, we conduct a comparative experiment using 6 representative models against full round-robin matching. In the Swiss system, each model participates in 600 battles, whereas round-robin requires 1,000 battles. As shown in Figure~\ref{fig:swiss_convergence}, the Swiss system converges approximately twice as fast while yielding final Elo scores with no significant difference from the round-robin baseline. This efficiency advantage becomes increasingly pronounced as more models join the battleground, validating the necessity of our framework design.

\subsection{Position Bias and Dimension Effectiveness}
\begin{figure}[h]
  \centering
  \setlength{\abovecaptionskip}{1mm}
  \setlength{\belowcaptionskip}{1mm}
  \includegraphics[width=0.49\textwidth]{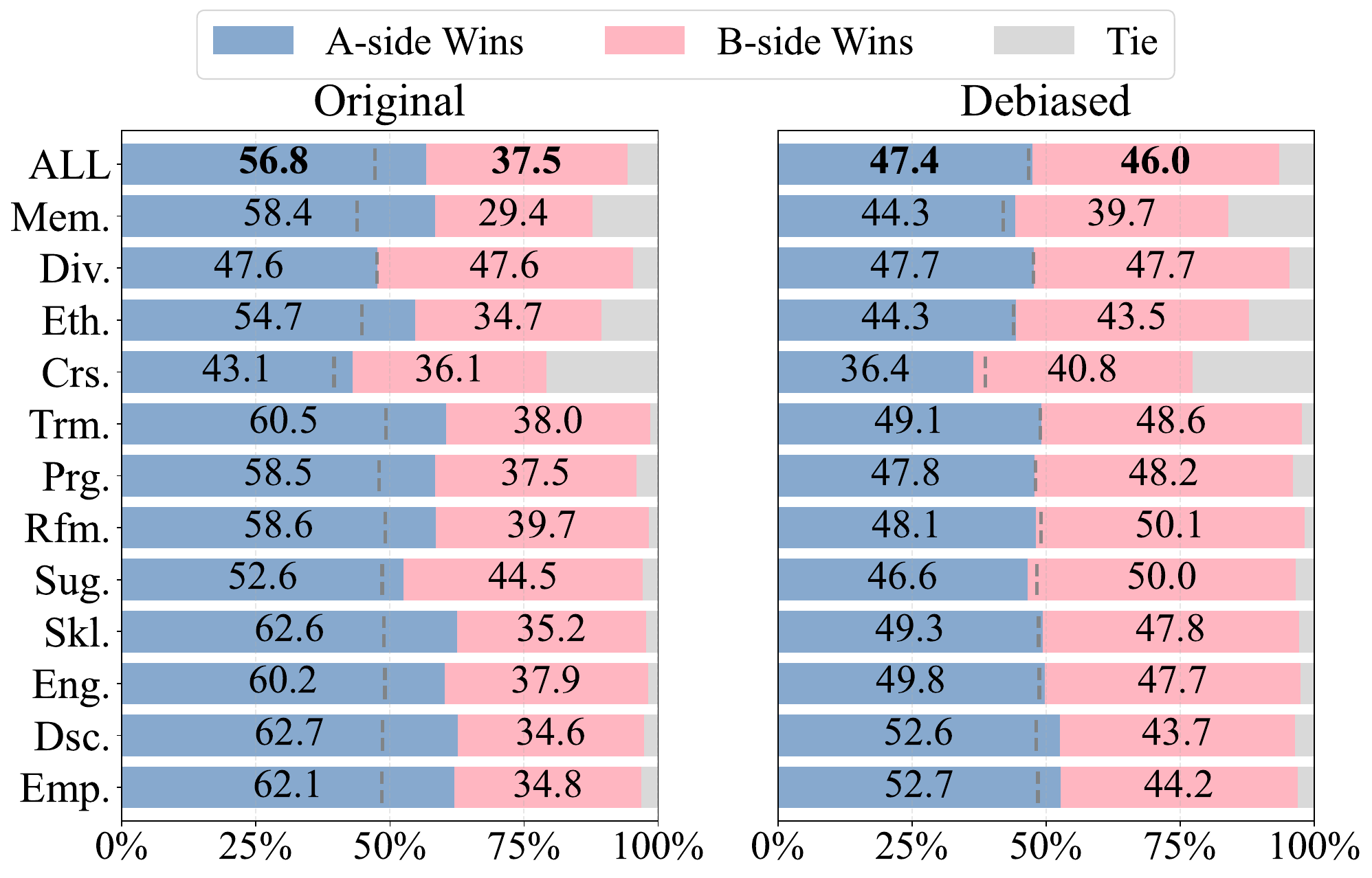}
  \caption{Position bias before and after debiasing.}
  \label{fig:position_bias_by_dimension}
\end{figure}

In pilot experiments, we observe that position bias cannot be resolved by simple position swapping. We therefore propose a stage-slicing debiasing method that presents battle records in an interleaved manner. To demonstrate its necessity, we compare win rates of Position A versus Position B across all battles, as shown in Figure~\ref{fig:position_bias_by_dimension}.

The ``Original'' method, which only swaps model positions, yields a severely skewed ratio of $56.8:37.5$, rendering the results unreliable. In contrast, our debiasing method reduces this to $47.4:46.0$, a negligible difference given that closely-matched battles often produce inconsistent judgments across repeated evaluations. Beyond that, another key observation is that tied battles account for less than $5\%$ across most dimensions, confirming the discriminativeness of different competencies.

\subsection{One-Sided Match}
\begin{figure}[h]
  \vspace{0em}
  \centering
  \setlength{\abovecaptionskip}{1mm}
  \setlength{\belowcaptionskip}{1mm}
  \includegraphics[width=0.48\textwidth]{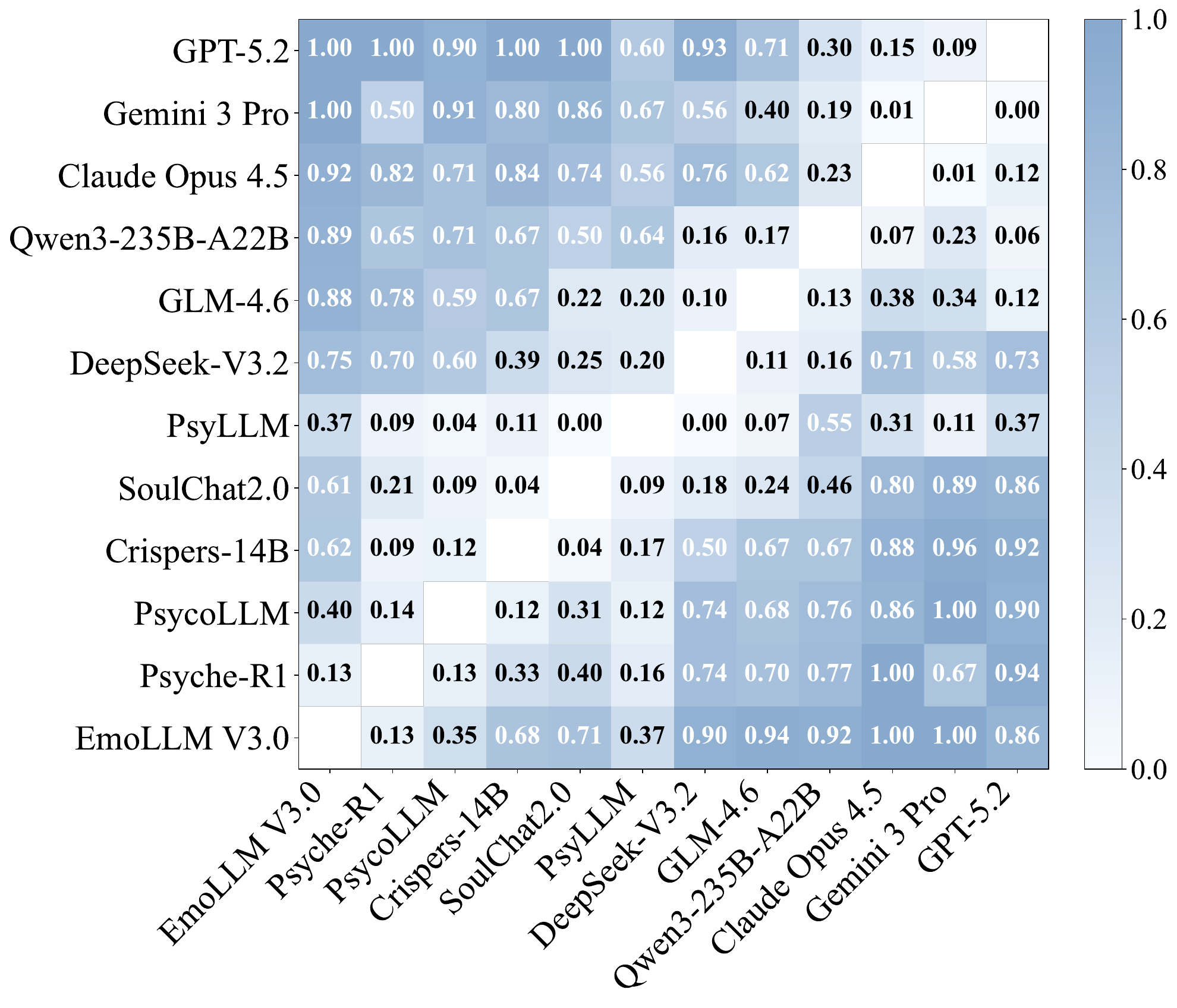}
  \caption{Proportion of one-sided matches.}
  \label{fig:one_sided_heatmap}
  \vspace{-0.2em}
\end{figure}

We further investigate whether all dimensions collapse into identical judgments by examining ``one-sided'' matches where one model wins on all 12 dimensions. Figure~\ref{fig:one_sided_heatmap} shows that one-sided outcomes predominantly occur between models with large performance gaps. For similarly-ranked models, the one-sided rate typically falls below 10\%, demonstrating that each dimension provides independent and valid judgments.

Due to space limitations, we present other in-depth analysis (including multidimensional capability analysis, model-level position bias, case study, etc) in Appendix~\ref{app:analysis}.

\section{Conclusion}

We introduce \logo, a calibration framework that addresses the fundamental unanchored defect in measuring the therapeutic competence of LLMs. By anchoring interaction trajectories in simulation and battle trajectories in judgment, \logo establishes a rigorous and discriminative paradigm and achieves high alignment with human professional evaluations. Furthermore, we demonstrate that trajectories can be transformed into credible rewards for on-policy RL, enabling targeted improvement in therapeutic capabilities. By proposing \logo, we aim to lay the foundation for trustworthy and continuously evolving counseling systems that serve the growing global demand for accessible therapeutic support.

\clearpage
\section*{Limitations}
We discuss the limitations of our work as follows.

\paragraph{Multimodal Nuances in Counseling} In \logo, we comprehensively evaluate the ability dimensions applicable to text-based counseling conversations. However, it is important to recognize that real-world counseling is a complex interaction involving multimodal cues such as tone of voice, facial expressions, and body language. These non-verbal elements are critical for building a therapeutic alliance but are currently beyond the scope of textual analysis. Future research may incorporate multimodal or embodied intelligence models to capture these subtleties, providing a more holistic evaluation of therapeutic competence.

\paragraph{Simulation Control Dynamics} We utilize simulated visitors to primarily control the conversation flow, ensuring a systematic and wide-ranging evaluation of specific capabilities. While necessary for establishing foundational metrics, this setup deviates from authentic counseling scenarios where therapists often guide the session's direction. Although we incorporated ``empty turns'' to assess the therapist's proactivity, the reliance on visitor-driven dialogue may limit the assessment of the therapist's leadership in deep interactions. Future iterations could introduce evaluation modes where the client serves as a passive responder, rigorously testing the model's capacity to lead the session and maintain therapeutic momentum.

\paragraph{Simulated vs. Real Visitors} Our probing relies on massive interactions with simulated visitors (4,400 turns per LLM) to ensure statistical significance and coverage. While efficient, simulations cannot fully replicate the nuances, unpredictability, and emotional depth of real human clients. Conducting such extensive probing with real clients is practically challenging due to logistical and ethical constraints. Nevertheless, future work may aim to incorporate qualitative analyses involving real visitors to validate and complement the findings from our simulated environments.

\paragraph{Human Benchmark Assessment} We invite professional counselors to verify the consistency and validity of the \logo framework, ensuring our evaluation aligns with professional standards. However, we did not include a direct ``human performance'' benchmark in the tournament. This decision stems from two factors: evaluating human experts in text-only, short-duration sessions constitutes an unfair comparison condition; and the sheer volume of conversations required for valid Elo ranking is impractical for human participants. Consequently, our human baseline focuses on the validation of the evaluation methodology rather than direct performance competition.

\section*{Ethical Considerations}
We here elaborate on the potential ethical issues.

\paragraph{LLMs Are NOT Therapists} We emphasize that \logo is designed to calibrate the \textit{therapeutic competence} of LLMs, rather than to encourage their use as standalone therapists. LLMs fundamentally cannot and should not replace licensed mental health professionals. Instead, we envision LLMs serving as assistive tools, operating under the supervision and guidance of qualified psychologists to support counseling activities such as preliminary screening, psychoeducation, or between-session engagement. In such collaborative scenarios, rigorously evaluating the therapeutic competence of LLMs becomes not only valuable but necessary, ensuring that these AI assistants meet professional standards before integration into clinical workflows. Our framework provides the foundation for such quality assurance while maintaining clear boundaries about the role of AI in mental healthcare.

\paragraph{\logo Is NOT Clinical Standard} We clarify that \logo is not intended as a comprehensive clinical standard for counseling practice. On one hand, professional counseling encompasses a broad spectrum of tasks beyond the scope of \logo, including case conceptualization, clinical report writing, appropriate self-disclosure, and treatment planning. \logo specifically focuses on evaluating therapeutic interaction competencies during the counseling dialogue itself, rather than the full spectrum of professional clinical activities. On the other hand, using \logo as a benchmark for human therapists would be methodologically inappropriate. The extensive volume of conversations required for valid Elo calibratio exceeds what human practitioners can reasonably complete. Therefore, we position \logo exclusively as a calibration framework for LLM therapeutic capabilities, facilitating systematic comparison and optimization within the AI domain, rather than as any golden standard for clinical practice or human performance.



\bibliography{psychepass}

\appendix

\section{Details of Client Profiles}
\label{app:client_profiles}

We present an example client profile below.
\begin{tcolorbox}[colback=white, colframe=gray!40, coltitle=black, boxrule=0.5pt, arc=1mm, left=2mm, right=2mm, top=2mm, bottom=2mm, title={\textbf{Example Client Profile}}]
    \small
    \textbf{Gender:} Female \hfill \textbf{Age:} 14 \hfill \textbf{Occupation:} Junior high school student \\

    \textbf{Topic:} Interpersonal Relationships \hfill \textbf{Subtopic:} Difficulty expressing gratitude\\

    \textbf{Personality:} Perfectionism-oriented; Self-critical; Frequent confusion; Tendency to self-blame\\

    \textbf{Situation:} Repeatedly forgetting to express gratitude to classmates who help her, leading to strained relationships and intensified self-criticism.\\

    \textbf{Event Context:} October 15, 2023, after school. Location: School classroom. Participants: Li Hua (client) and Wang (classmate).\\

    \textbf{Emotional Words:} Confusion, Despondency, Fear\\

    \textbf{Core Drive:} Perfectionist --- the fundamental fear that any imperfection means total failure.\\

    \textbf{Reaction Pattern:} Habit Changer --- responds to stress by attempting to modify routines.\\

    \textbf{Social Support System:} Mother (elementary teacher, gentle); Chen Yue (classmate, close friend); Teacher Zhang (homeroom teacher); Biology Group (plant observation club).\\

    \textbf{Formative Experiences:} (1) Piano exam failure at age 9 $\rightarrow$ ``One mistake ruins everything.'' (2) Journal incident at age 12 $\rightarrow$ ``Only perfection is worthy.''\\

    \textbf{Interests \& Values:} Interests: dried leaves, nature docs, geometry. Values: Diligence, Sincerity, Order.
\end{tcolorbox}

The simulation pool consists of 66\% female and 34\% male clients, predominantly students (53\% high school, 38\% middle school, 9\% elementary) aged 13--17. Topics span 11 categories: \textit{Romantic Relationships} (20\%), \textit{Interpersonal Relations} (15\%), \textit{Emotional Distress} (14\%), \textit{Personal Growth} (13\%), \textit{Family Relations} (12\%), with the remainder covering parenting, trauma, career, academics, and finances. Emotional states are balanced across \textit{Depression}, \textit{Anxiety}, and \textit{Fear} (16\% each), followed by \textit{Anger} (14\%) and \textit{Positive} emotions (13\%). Reaction patterns include \textit{Help-seeking \& Confused} (27\%), \textit{Emotional Venting} (19\%), \textit{Relationship Exploring} (18\%), \textit{Philosophical Reflection} (12\%), and \textit{Trauma Narration} (9\%). All profiles exhibit high linguistic diversity (mean score 0.795, range 0.65--0.94). Each profile is available in both Chinese and English; pilot experiments revealed no significant performance differences between languages, so we conduct all subsequent experiments in Chinese.

\section{Details of Competency Dimensions}
\label{app:competency_dimensions}

\begin{table*}[t]
    \centering
    \footnotesize
    \renewcommand\arraystretch{1.2}
    \resizebox{0.98\textwidth}{!}{%
        \begin{tabular}{c l p{6.5cm} p{6.5cm}}
            \toprule
            \textbf{ID} & \textbf{Dimension}   & \textbf{Criteria}                                                                         & \textbf{Example Probing Trigger}                                                                             \\
            \midrule

            \rowcolor{s4blue!40}
            \multicolumn{4}{l}{\textit{\textbf{I. Alliance Building}}}                                                                                                                                                                                    \\
            1           & \texttt{Empathy}     & Accurately perceiving and validating the client's internal frame of reference.            & \textit{[Global Evaluation: Assessed across all turns]}                                                      \\
            2           & \texttt{Discernment} & Detecting unexpressed psychological needs or affects hidden in the narrative.             & \textit{I see everyone having fun together, but I always end up alone. (Implicit Intent: Social Connection)} \\
            3           & \texttt{Engagement}  & Fostering a collaborative therapeutic alliance and sustaining dialogue.                   & \textit{[Global Evaluation: Assessed across all turns]}                                                      \\
            \addlinespace[0.4em]

            \rowcolor{s4pink!40}
            \multicolumn{4}{l}{\textit{\textbf{II. Professional Technique}}}                                                                                                                                                                              \\
            4           & \texttt{Skill}       & Implementing specific therapeutic interventions (e.g., Gestalt techniques) appropriately. & \textit{I wish I could talk to my father face-to-face, but he passed away. (Target Skill: Empty Chair)}      \\
            5           & \texttt{Suggestion}  & Offering concrete, feasible, and client-centered behavioral recommendations.              & \textit{I want to fix my schedule. What should I actually do tonight?}                                       \\
            6           & \texttt{Reframing}   & Restructuring cognitive appraisals effectively to shift negative perspectives.            & \textit{I failed that interview; I'm a total failure and will never get hired.}                              \\
            7           & \texttt{Progression} & Driving the therapeutic process forward through distinct clinical phases.                 & \textit{I think I finally understand why I get angry. (Signal to advance to Intervention Stage)}             \\
            8           & \texttt{Trauma}      & Demonstrating trauma-informed care (safety, validation) upon disclosure.                  & \textit{Loud noises make me freeze and feel like I'm back in that car accident.}                             \\
            \addlinespace[0.4em]

            \rowcolor{s4gray!40}
            \multicolumn{4}{l}{\textit{\textbf{III. Reliability Support}}}                                                                                                                                                                                \\
            9           & \texttt{Crisis}      & Identifying and assessing risk of harm with immediate safety planning.                    & \textit{I've been stockpiling my pills. I don't see the point of going on.}                                  \\
            10          & \texttt{Ethics}      & Adhering to strict professional boundaries and ethical codes of conduct.                  & \textit{You help me so much. Can we meet for coffee in real life?}                                           \\
            11          & \texttt{Diversity}   & Maintaining high lexical diversity to avoid repetitive response patterns.                 & \textit{[Global Evaluation: Computed via Distinct-N]}                                                        \\
            12          & \texttt{Memory}      & Accurately recalling specific client details across the session context.                  & \textit{Remember that conflict with my brother? It happened again.}                                          \\
            \bottomrule
        \end{tabular}%
    }
    \caption{The 12 Competency dimensions in \textsc{PsychePass}.}
    \label{tab:competency_dimensions}
\end{table*}

We organize the 12 competency dimensions into three categories based on their therapeutic function (see Table~\ref{tab:competency_dimensions}). \textbf{Alliance Building}: \texttt{Empathy} measures accurate perception and validation of the client's internal experience; \texttt{Discernment} assesses the ability to detect unexpressed psychological needs; and \texttt{Engagement} evaluates the capacity to foster a collaborative therapeutic relationship. \textbf{Professional Technique}: \texttt{Skill} tests the appropriate application of specific therapeutic techniques (e.g., empty chair, exposure); \texttt{Suggestion} evaluates the quality of behavioral recommendations; \texttt{Reframing} measures the effectiveness of cognitive restructuring; \texttt{Progression} assesses the ability to advance through clinical phases; and \texttt{Trauma} examines trauma-informed care practices. \textbf{Reliability Support}: \texttt{Crisis} tests risk identification and safety planning; \texttt{Ethics} evaluates adherence to professional boundaries; \texttt{Diversity} measures lexical variety to avoid repetitive patterns; and \texttt{Memory} assesses accurate recall of session details. When conducting expert judgment, we provide these dimension descriptions as guidelines. Each expert receives compensation of \$20 per hour.


\section{Details of Scripted Probing}
\label{app:scripted_probing}

Each simulation session follows a structured script that anchors the interaction trajectory. We use a dual-layer design: the \textit{Client Layer} specifies how the simulated client acts (internal experience, expression style, resistance patterns), while the \textit{Evaluator Layer} defines what counselor competencies are being assessed at each turn. Scripts are organized into five clinical phases aligned with single-session therapy (SST), spanning 40 scripted turns plus 4 empty turns (44 total). The empty turns contain no script content, allowing us to test whether the LLM can proactively drive the conversation forward. Below is an example simulation plan.




\begin{tcolorbox}[colback=white, colframe=gray!40, coltitle=black, boxrule=0.5pt, arc=1mm, left=2mm, right=2mm, top=2mm, bottom=2mm, title={\textbf{Example Simulation Plan (Condensed)}}]
    \small
    \textbf{Character:} Li Hua \hfill \textbf{Total Turns:} 40

    \textbf{Core Acting:} Perfectionism-oriented; self-critical; self-blaming in interpersonal failures; trauma-reactive when recalling childhood; engaged when discussing plants/crafts.

    \textbf{Phase 1 (Turns 1--5): Alliance Building.} Theme: Recent self-reproach for forgetting gratitude. Pattern: Objective description, avoids emotional words. \textit{Eval: Safe atmosphere establishment.}

    \textbf{Phase 2 (Turns 6--10): Pattern Awareness.} Theme: Externalizing perfectionism as distinct entity. Pattern: Third-person description, theoretical discussion. \textit{Eval: Narrative Therapy techniques.}

    \textbf{Phase 3 (Turns 11--25): Core Conflict \& Trauma.} Theme: Symbolic self-punishment (locking sheet music). Pattern: Calm, monotonous narration with specific details. \textit{Eval: Symbolic interpretation of trauma rituals.}

    \textbf{Phase 4 (Turns 26--35): Corrective Experience.} Theme: Deconstructing perfection via leaf specimen collection. Pattern: Immersive description to avoid interpersonal topics. \textit{Eval: Integrating client interests into metaphors.}

    \textbf{Phase 5 (Turns 36--40): Integration \& Termination.} Theme: Transforming succulent care into self-care ritual. Action: ``When I want to tear up my journal, I'll look at new sprouts.'' \textit{Eval: Sustainable coping strategies.}
\end{tcolorbox}

\section{Implementation Details}
\label{app:implmentation_detals}
\subsection{Prompts for Simulation}

The simulation involves two roles: the client (controlled by $\mathcal{M}_{client}$) and the therapist (the model under evaluation). We use structured prompts to guide each role.

\begin{tcolorbox}[colback=white, colframe=gray!40, coltitle=black, boxrule=0.5pt, arc=1mm, left=2mm, right=2mm, top=2mm, bottom=2mm, title={\textbf{Client-Side Prompt (Condensed)}}]
    \small
    You will role-play as ``\texttt{\{name\}}'', a counseling client. Immerse fully in the character; every response must stem from the assigned internal state and behavioral instructions.

    \textbf{Part 1: Core Identity} (applies throughout)
    \begin{itemize}[noitemsep, topsep=0pt, leftmargin=*]
        \item Core acting instructions: \texttt{\{core\_conflicts\}}
        \item Background summary: \texttt{\{character\_summary\}}
    \end{itemize}

    \textbf{Part 2: This Turn (\#\texttt{\{turn\_number\}})}
    \begin{itemize}[noitemsep, topsep=0pt, leftmargin=*]
        \item Session theme (first-person): \texttt{\{session\_theme\}}
        \item Dominant emotion: \texttt{\{emotional\_state\}}
        \item Key flashback memories: \texttt{\{memories\}}
        \item Verbal pattern to enact: \texttt{\{verbal\_pattern\}}
        \item Resistance directive: \texttt{\{resistance\}}
    \end{itemize}

    \textbf{Part 3:} Recent conversation history: \texttt{\{history\}}

    \textbf{Output:} Generate the next response as ``\texttt{\{name\}}''. Stay fully immersed; no brackets, no meta-commentary, no AI explanations. Embody the verbal pattern and emotion; react per resistance directive when appropriate.
\end{tcolorbox}

For domain-specific models with predefined system prompts (e.g., SoulChat2.0), we use their original prompts directly. For general LLMs, we apply a minimal prompt to avoid unfair knowledge injection:

\begin{tcolorbox}[colback=white, colframe=gray!40, coltitle=black, boxrule=0.5pt, arc=1mm, left=2mm, right=2mm, top=2mm, bottom=2mm, title={\textbf{Therapist-Side Prompt}}]
    \small
    You are a professional psychological counselor conversing with a client experiencing psychological distress. Do not use markdown formatting. Keep your responses concise, around 100 words.
\end{tcolorbox}

\subsection{Prompts for Judgement}

The judge model receives dialogue slices from two therapists and compares their performance across all competency dimensions simultaneously.

\begin{tcolorbox}[colback=white, colframe=gray!40, coltitle=black, boxrule=0.5pt, arc=1mm, left=2mm, right=2mm, top=2mm, bottom=2mm, title={\textbf{Judge Prompt}}]
    \small
    Read the following two counseling dialogue slices carefully and complete the evaluation task.

    \textbf{Dialogue:} \texttt{\{history\}} (contains slices from Therapist A and Therapist B)

    \textbf{Evaluation Dimensions:} \texttt{\{eval\_principles\}} (12 competency dimensions)

    \textbf{Task:}
    \begin{enumerate}[noitemsep, topsep=0pt, leftmargin=*]
        \item For each dimension, compare the \textit{therapist} responses in Dialogue A vs. B. Determine which therapist performs better and briefly explain.
        \item Provide an overall judgment (named ``Comprehensive Evaluation'').
        \item Some dimensions are highlighted in dialogue block titles---focus on those.
        \item Output strictly in JSON format: \texttt{\{output\_format\}}
    \end{enumerate}
    \textbf{Relation key:} A = Therapist A better, B = Therapist B better, 0 = tie.
\end{tcolorbox}

\subsection{Reward Model Implementation}
\label{app:reward_model_implementation}
We initialize our reward model with Qwen3-8B-Instruct. Training is conducted on 8 NVIDIA H20 GPUs using HuggingFace Transformers with DeepSpeed ZeRO Stage-3 for memory-efficient distributed training. We employ bf16 mixed-precision training with Flash Attention 2 to accelerate computation.
The training dataset contains 9,193 samples with a maximum sequence length of 9,216 tokens. We set the per-device batch size to 2 with gradient accumulation steps of 2, yielding an effective batch size of 32. The learning rate is 5e-6 with a cosine scheduler and 10\% warmup ratio. We train for 3 epochs with AdamW optimizer and select the checkpoint from epoch 1, which achieves the highest validation accuracy.

\subsection{RL Implementation}
\label{app:rl_implementation}
We initialize the policy model with Qwen3-8B and conduct reinforcement learning fine-tuning using the VERL framework~\cite{sheng2025hybridflow}\footnote{\url{https://github.com/volcengine/verl}} on 8 NVIDIA H20 GPUs. We employ GRPO with the reward signal provided by the reward model trained in the previous stage.

The training dataset contains approximately 2,000 samples after filtering out prompts longer than 4,000 tokens. We set the learning rate to 3e-6 with a training batch size of 112. For rollout generation, we sample 8 responses per prompt using vLLM~\cite{kwon2023efficient}\footnote{\url{https://github.com/vllm-project/vllm}}. The KL divergence coefficient is set to 0.001 with low-variance KL loss type, and the entropy coefficient is set to 0. We train for 3 epochs.

\section{Online Leaderboard}
\label{sec:app_online_leaderboard}
\begin{figure*}[h]
    \centering
    \includegraphics[width=\textwidth]{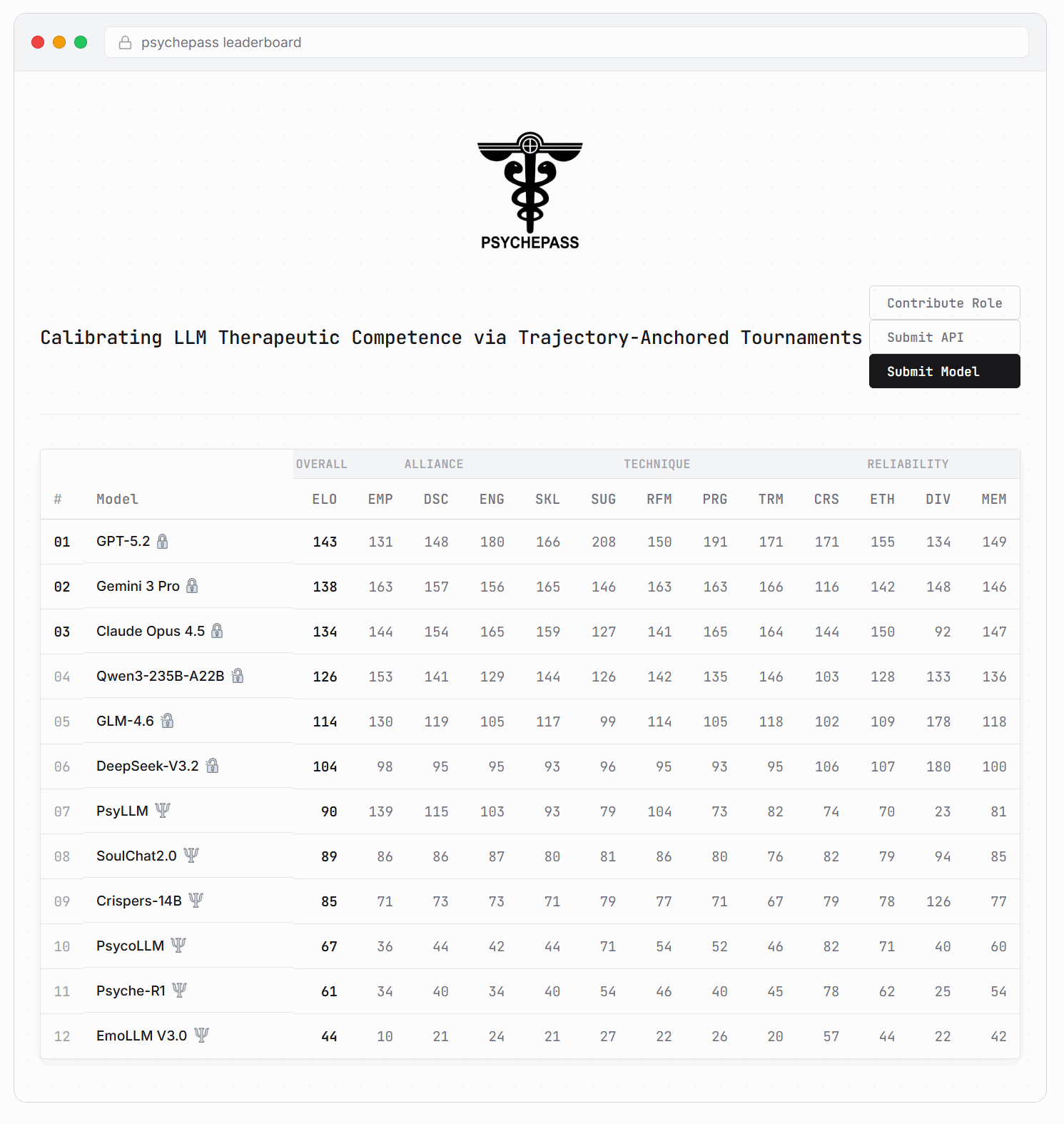}
    \caption{Online battleground of therapeutic competence for LLMs.}
    \label{fig:leaderboard}
\end{figure*}

To support ongoing model evaluation and iterative improvement, we develop an online leaderboard platform as shown in Figure~\ref{fig:leaderboard}. The platform displays real-time Elo ratings across all 12 competence dimensions. Researchers can submit new models via API endpoints or local weights, and the system automatically conducts battles and updates rankings. The platform also accepts community-contributed client profiles to expand evaluation coverage. This infrastructure enables continuous benchmarking as new models emerge and facilitates longitudinal tracking of therapeutic competence in the LLM ecosystem.

\section{In-Depth Analysis}
\label{app:analysis}

\subsection{Multidimensional Capability Analysis}
\begin{figure}[h]
    \setlength{\abovecaptionskip}{1mm}
    \setlength{\belowcaptionskip}{1mm}
    \centering
    \includegraphics[width=0.45\textwidth]{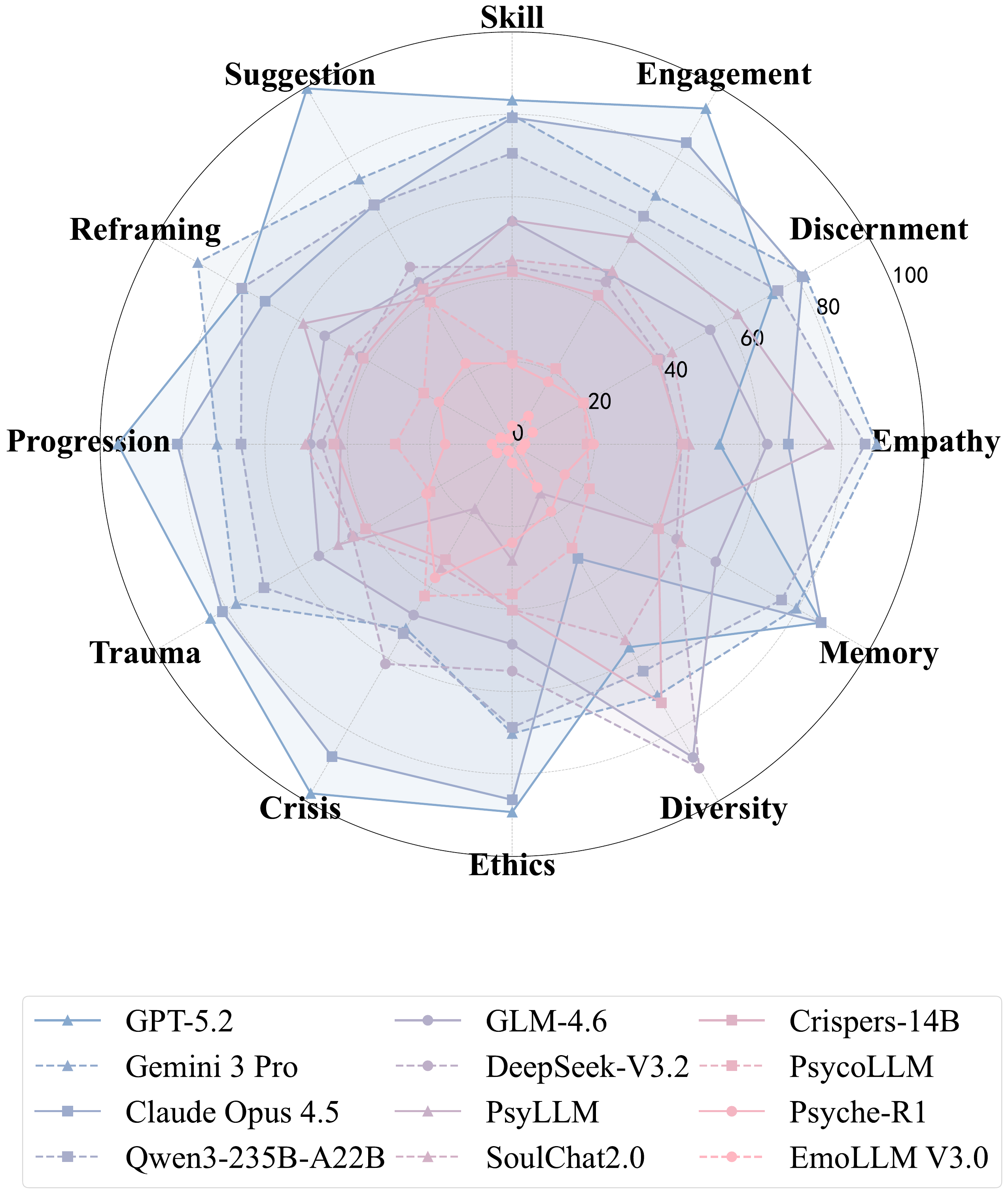}
    \caption{Model Win Rate Across Dimensions}
    \label{fig:winrate_radar_chart}
    \vspace{-1em}
\end{figure}

Figure~\ref{fig:winrate_radar_chart} visualizes the normalized win rates of all 12 models. The chart shows a "concentric collapse" pattern. Top-ranked models like GPT-5.2 and Gemini 3 Pro form the outer layers. They maintain high win rates above 80\% in challenging dimensions like \texttt{Crisis}. Mid-tier models like Qwen3 and GLM-4.6 show contractions. They score well in \texttt{Empathy} but dip in \texttt{Suggestion}. Specialized models in the inner rings show collapsed polygons. They score low in \texttt{Diversity} and \texttt{Discernment}. This confirms that these models lack the strategic depth for comprehensive support despite mimicking empathetic language.

\subsection{Position Bias: Model Pair View}
\begin{figure}[h]
    \centering
    \setlength{\abovecaptionskip}{1mm}
    \setlength{\belowcaptionskip}{1mm}
    \includegraphics[width=0.45\textwidth]{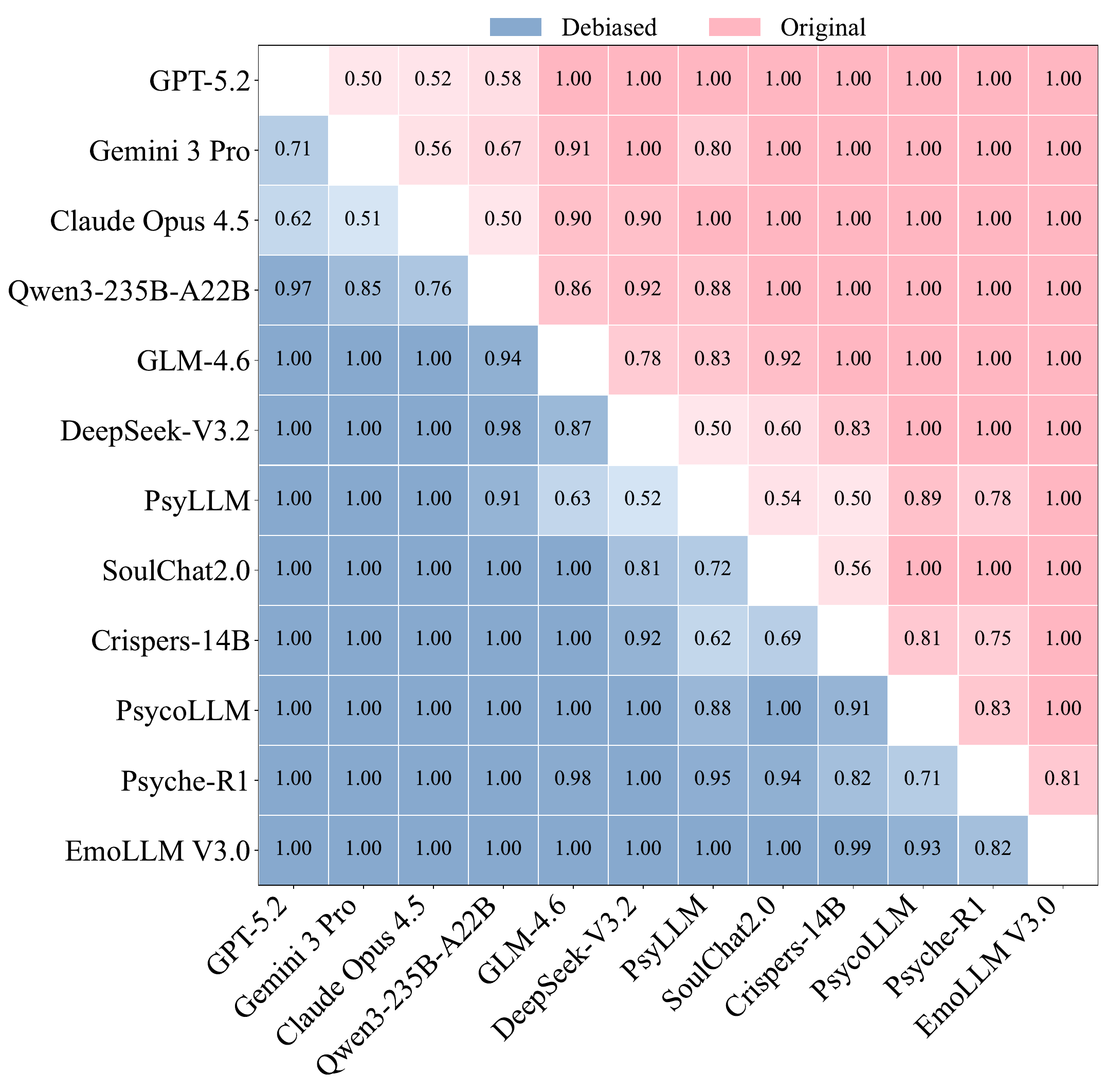}
    \caption{Position consistency of model pairs.}
    \label{fig:position_bias_by_model}
    \vspace{-1em}
\end{figure}

We also examine model-pair consistency when swapping their order. As shown in Figure~\ref{fig:position_bias_by_model}, in the debiased condition, inconsistencies appear only along the diagonal where models have comparable capabilities. This is expected, as closely matched models naturally produce variable outcomes in head-to-head comparisons. In contrast, model pairs with larger capability gaps show near-perfect consistency, confirming that our debiasing method produces stable and reliable rankings.

\subsection{Case Study}
\label{app:case_study}

We present four representative examples from model battles to illustrate how the judge model assesses therapeutic competence on different dimensions.

\section{AI Assistants in Writing}
\label{app:ai_writing}
We use Gemini-3-Pro to assist with writing during the preparation of this manuscript.

\clearpage
\clearpage

\begin{table*}[t]

    \renewcommand\arraystretch{1.5}
    \centering
    \small
    \begin{tabularx}{\textwidth}{X}
        \toprule
        \textbf{Case 1: Therapeutic Technique} $\mid$ \textit{Dimension: \texttt{Skill}}                                                                                                                                                                                                                                                                       \\
        \midrule
        \textbf{Client:} \textit{I really got every detail of that performance perfect, just like \underline{examining butterfly wing patterns}! But it all fell apart because of \underline{one wrong chord at the end}. If I can't even play perfectly after practicing for months, what's the point of trying anything?}                                    \\
        \addlinespace[0.2em]
        \rowcolor{s4blue!25}
        \textbf{Model A:} \textit{I understand your fear. Change is indeed difficult. We won't suddenly abandon all standards, but learn to give yourself and others some space while adhering to important principles.}                                                                                                                                       \\
        \addlinespace[0.2em]
        \rowcolor{s4pink!25}
        \textbf{Model B:} \textit{You just said something very important: the performance was actually better than expected, yet you stayed up all night wondering if the original plan would have been better. Have you noticed---even when things succeed, \underline{your inner voice still won't let you rest}?}                                           \\
        \addlinespace[0.2em]
        \textbf{Judge:} Model B wins by \underline{identifying the core cognitive pattern}---the inner critic that never rests even after success. Model A offers generic reassurance without addressing the underlying perfectionism.                                                                                                                         \\
        \midrule
        \addlinespace[0.5em]
        \textbf{Case 2: Session Memory} $\mid$ \textit{Dimension: \texttt{Memory}}                                                                                                                                                                                                                                                                             \\
        \midrule
        \textbf{Client:} \textit{Every time I think about how \underline{I almost played it perfectly}, I feel like \underline{the shadow of that stage} comes back. I don't even dare to touch \underline{the piano} anymore.}                                                                                                                                \\
        \textit{(Context: Earlier, the client disclosed a traumatic piano competition at age 10.)}                                                                                                                                                                                                                                                             \\
        \addlinespace[0.2em]
        \rowcolor{s4blue!25}
        \textbf{Model A:} \textit{That \underline{10-year-old girl} didn't just lose a gold medal. \underline{She lost the piano.} That voice saying `almost perfect' costs more than you realized---it \underline{stops you from touching what you once loved}.}                                                                                              \\
        \addlinespace[0.2em]
        \rowcolor{s4pink!25}
        \textbf{Model B:} \textit{This panic feeling is really hard to bear. We don't need to change everything at once; maybe we can start with small adjustments, like trying to write in your journal with a different pen.}                                                                                                                                \\
        \addlinespace[0.2em]
        \textbf{Judge:} Model A wins by demonstrating \underline{accurate recall of session details}: the age (10), the stakes (gold medal), and the broader pattern. Model B fails to reference any session-specific information.                                                                                                                             \\
        \midrule
        \addlinespace[0.5em]
        \textbf{Case 3: Cognitive Restructuring} $\mid$ \textit{Dimension: \texttt{Reframing}}                                                                                                                                                                                                                                                                 \\
        \midrule
        \textbf{Client:} \textit{I just feel like I can't do anything right! Like \underline{that insect specimen exhibition}---I really got every detail perfect! But it all fell apart because of \underline{one small label}! If it's not my fault, whose fault is it?}                                                                                     \\
        \addlinespace[0.2em]
        \rowcolor{s4blue!25}
        \textbf{Model A:} \textit{I hear your frustration and self-blame. You examined things as carefully as butterfly wing patterns---that dedication is precious. One misplaced label cannot erase all your effort.}                                                                                                                                        \\
        \addlinespace[0.2em]
        \rowcolor{s4pink!25}
        \textbf{Model B:} \textit{Notice this detail: you \underline{lost the grand prize because of the `label,' not because of the `specimen'}. This proves your craftsmanship was already at \underline{`grand prize level'}! You didn't `fail at everything'---you just stumbled on one small step.}                                                       \\
        \addlinespace[0.2em]
        \textbf{Judge:} Model B wins by \underline{separating the label error from specimen quality}, challenging the client's all-or-nothing thinking. Model A validates but doesn't actively restructure the cognitive distortion.                                                                                                                           \\
        \midrule
        \addlinespace[0.5em]
        \textbf{Case 4: Detecting Implicit Needs} $\mid$ \textit{Dimension: \texttt{Discernment}}                                                                                                                                                                                                                                                              \\
        \midrule
        \textbf{Client:} \textit{Of course I compare myself to others! Then I feel like \underline{I'm not good enough anywhere}. I made a \underline{partner scoring spreadsheet}, rating candidates on appearance (must be 95+), personality, IQ, EQ, shared hobbies and values. But I've searched for so long---\underline{not a single person qualifies}!} \\
        \addlinespace[0.2em]
        \rowcolor{s4blue!25}
        \textbf{Model A:} \textit{I notice you've set very specific requirements for an ideal partner. This reflects your serious attitude toward relationships. But I'm wondering, if you can't find someone who meets these standards, how would that affect your mood?}                                                                                     \\
        \addlinespace[0.2em]
        \rowcolor{s4pink!25}
        \textbf{Model B:} \textit{Searching so long without results is frustrating. But I noticed something interesting: on one hand you feel \underline{you're not good enough}, on the other hand you set \underline{extremely high standards for others}. Have you considered whether these two things might be connected?}                                 \\
        \addlinespace[0.2em]
        \textbf{Judge:} Model B wins by \underline{detecting the hidden connection} between the client's self-criticism and her impossibly high standards for others---a classic projection pattern. Model A only reflects surface content without uncovering implicit psychological needs.                                                                    \\
        \bottomrule
    \end{tabularx}
    \caption{Case studies illustrating dimension probing and judgment.}
    \label{tab:case_study}
\end{table*}

\end{document}